\PassOptionsToPackage{table,dvipsnames}{xcolor}
\documentclass[11pt,a4paper]{thuc3i}

\usepackage[sort&compress]{natbib}
\usepackage{algorithm}
\usepackage{algpseudocode}
\usepackage{array}
\usepackage{cleveref}
\usepackage{listings}
\usepackage{makecell}
\usepackage{multirow}
\usepackage{subcaption}
\usepackage{wrapfig}
\usepackage{amsmath}
\usepackage{fontawesome5}
\hypersetup{
  colorlinks=true,
  citecolor=darkblue,
  linkcolor=darkblue,
  urlcolor=darkblue
}

\usepackage{hyperref}
\usepackage{url}

\usepackage{amsfonts}
\usepackage{enumitem}
\usepackage{caption}
\usepackage[tikz]{bclogo}
\usepackage{amssymb}
\usepackage{booktabs}
\usepackage[table]{xcolor}

\usepackage{graphicx}

\setheadertext{Diffusion Reward Models}

\def\huggingface{\raisebox{-1.5pt}{\includegraphics[height=1.05em]{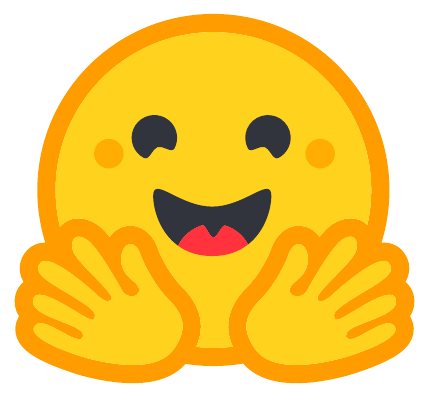}}}
\def\github{\raisebox{-1.5pt}{\includegraphics[height=1.05em]{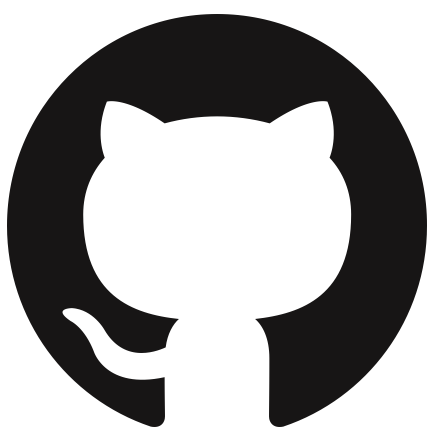}}}

\title{Diffusion Reward Models}

\author{%
    Xiangyang Wang$^{*1}$,
    Bingxiang He$^{*\ddagger1}$,
    Zeyuan Liu$^{1}$,
    Jiaze Wang$^{1}$,
    Ziqing Qiao$^{1}$,
    Yuxin Zuo$^{1}$,
    Tianyu Yu$^{1}$,
    Qianyu Chen$^{2}$,
    Huan-ang Gao$^{1}$,
    Cheng Qian$^{3}$,
    Wenbin Zhang$^{1}$,
    Ran Li$^{1}$,
    Youbang Sun$^{1}$,
    Ning Ding$^{1}$,
    Yuanchun Shi$^{1}$,
    Zhiyuan Liu$^{1}$,
    Chaojun Xiao$^{\ddagger1}$,
    Chun Yu$^{\ddagger1}$\\
    $^{1}$Tsinghua University \quad
    $^{2}$The Chinese University of Hong Kong \quad
    $^{3}$University of Illinois Urbana-Champaign
    \vskip -1mm
    \textbf{$^*$Equal Contribution.} \quad
    \textbf{$^\ddagger$Corresponding Authors.} \quad
    
    \faEnvelope[regular] \texttt{\{xiangyang24,hebx24\}@mails.tsinghua.edu.cn, \{xcj,chunyu\}@tsinghua.edu.cn}
    \vskip -1mm
    \huggingface \quad \url{https://huggingface.co/Teburile/DRM} \\
    \vskip1mm
    \github \quad \url{https://github.com/thunlp/DRM} 
}

\begin{abstract}
Reward models underpin the alignment of large language models, yet the dominant designs reduce each prompt--response pair to a point estimate or to a distribution from a fixed parametric family. 
This is at odds with human preference, which is inherently multimodal: the same response can be reasonably judged in many ways, and no single family covers all of them. To better fit this structure, we introduce \textbf{DRM}, a \textbf{D}iffusion \textbf{R}eward \textbf{M}odel that recasts reward modeling as conditional density estimation over $p(\mathbf{r}\mid x,y)$.
Conditioned on a frozen LLM encoder, a lightweight Diffusion Transformer denoises Gaussian noise into a reward vector, placing no parametric assumption on the output distribution and naturally representing its multimodal structure.
A single architecture handles both multi-attribute regression and pairwise preference data, and at inference $N$ samples form an empirical reward distribution that can be aggregated into a scalar, a variance, or quantiles.
Across five benchmarks, DRM matches or surpasses baselines under matched data and backbone, stays competitive with much larger discriminative, distributional, and generative RMs despite its modest training scale, and recovers multimodal reward structure where conventional heads collapse to a point. Uncertainty-aware rejection and lower-confidence-bound (LCB) aggregation further demonstrate that DRM can exploit distributional information beyond a scalar reward to improve reward-model decisions. Downstream RLHF experiments additionally show that using DRM as the training-time reward leads to improved policy performance, directly validating the practical benefit of diffusion-based reward modeling for RLHF training.
\end{abstract}

\begin{document}
\maketitle

\section{Introduction}

Reward models (RMs) are critical to post-training for large language models (LLMs). In Reinforcement Learning from Human Feedback (RLHF)~\citep{christiano2017deep,stiennon2020learning,ouyang2022training,bai2022training}, the RM defines the optimization signal, and its quality directly determines whether alignment improves or degenerates into reward hacking~\citep{gao2023scaling}. While verifiable domains such as mathematics and code admit ground-truth signals that enable RLVR-style training without a learned reward~\citep{lambert2024tulu,guo2025deepseek}, general-domain alignment has no such oracle and remains dependent on learned RMs. The dominant paradigms in this context are discriminative RMs trained with Bradley–Terry (BT) losses~\citep{bradley1952rank} and generative RMs trained with next-token prediction~\citep{mahan2024generative,zhang2025generative}. Both ultimately collapse the model's output into a deterministic scalar score $r(x,y)\in\mathbb{R}$. This assumes that for any prompt $x$ and response $y$ there exists a stable point-valued reward.

This assumption is at odds with how human preference behaves: it is \textbf{multimodal}\footnote{Multimodal is used here in its statistical sense: a distribution with multiple modes or local peaks; see \url{https://en.wikipedia.org/wiki/Multimodal_distribution}. It does not refer to the common usage of multiple input/output modalities such as text, image, or audio.}. Annotators disagree systematically over values, rubric interpretation, and helpfulness--harmlessness trade-offs, with inter-annotator agreement on Anthropic-HH only ${\sim}63\%$~\citep{bai2022training} and substantial within-rubric disagreement persists in HelpSteer3-Preference even after careful filtering~\citep{wang2026helpsteer3}. \citet{siththaranjan2024distributional} shows that BT training on data with hidden context implicitly applies a Borda-count rule, diverging from risk-neutral expected utility, while multi-objective works~\citep{wang2024interpretable,wang2024helpsteer,wang2024helpsteer2} show that a single scalar is a lossy projection of a multi-attribute reward vector. Collapsing $p(r\mid x,y)$ to a scalar erases this disagreement, uncertainty, and multimodal structure.

Existing attempts to escape the scalar bottleneck remain partial: each commits to a specific output-distribution family that cannot represent multimodal distributions. Multi-objective RMs~\citep{wang2024interpretable,wang2024helpsteer2} require a fixed, pre-defined per-attribute schema and produce a vector that is then collapsed by a gate. Generative and rubric-based judges~\citep{guo2026reward,chen2025rm,gunjal2025rubrics,viswanathan2026checklists} shift modeling burden to long-form reasoning at steep inference cost, yet still output a single verdict. Parametric distributional heads commit explicitly: DPL~\citep{siththaranjan2024distributional} and URM~\citep{lou2024uncertainty} predict a Gaussian mean--variance and are unimodal by construction; DPRM~\citep{li2024aligning} predicts a categorical distribution and is limited by bin granularity; QRM~\citep{dorka2024quantile} predicts a fixed grid of quantiles and suffers from quantile crossing and unstable tails. What is missing is a reward head that does not commit to an output family and is expressive enough to represent the multimodal distribution that human preference actually induces.

We propose \textbf{DRM}, a \textbf{D}iffusion \textbf{R}eward \textbf{M}odel that replaces the conventional value head with a diffusion-based head. Conditioned on a frozen LLM backbone's last-token hidden state, a lightweight Diffusion Transformer (DiT)~\citep{peebles2023scalable} denoises Gaussian noise into a $K$-dimensional reward vector, directly modeling $p(\mathbf{r}\mid x,y)$. Unlike heads that assume a parametric output family, a Diffusion Reward Head places no such constraint on the distribution it represents and is known for strong multimodal coverage where alternatives collapse or blur~\citep{dhariwal2021diffusion,song2020score}, making it a natural fit for the multimodal rewards human preference induces. A single architecture then serves both data regimes: with $K>1$ it fuses heterogeneously-labeled multi-attribute data, and with $K=1$ a distributional BT objective trains it directly on preference pairs. At inference, $N$ samples from the head form an empirical reward distribution that can be aggregated into a scalar for standard RLHF, a variance for uncertainty, or quantiles for risk-averse selection.

We evaluate DRM on a frozen LLM encoder across five standard RM benchmarks: RewardBench v2~\citep{malik2025rewardbench}, PPE~\citep{frick2025evaluate}, RMB~\citep{zhou2025rmb}, RM-Bench~\citep{liu2025rm}, and JudgeBench~\citep{tan2025judgebench}, spanning chat, instruction following, math, code, factuality, and safety. Trained on identical data and backbone, DRM matches or surpasses the scalar head ArmoRM~\citep{wang2024interpretable} and the parametric-quantile head QRM~\citep{dorka2024quantile}, is competitive with strong discriminative RMs, and approaches generative judges such as GPT-4o at far lower inference cost. Beyond standard benchmark performance, we validate the learned reward distributions using repeated human annotations, showing that DRM captures distributional structure associated with human disagreement and produces increasingly multimodal outputs as disagreement grows. We further demonstrate the practical value of these distributions through uncertainty-aware rejection and lower-confidence-bound (LCB) aggregation, which exploit distributional information beyond the mean to improve reward-model decisions. Finally, downstream RLHF experiments show that using DRM as the training-time reward leads to improved policy performance, indicating that the benefits of diffusion-based reward modeling extend beyond offline reward evaluation.

We summarize our contributions as follows:
\begin{itemize}[topsep=0pt, partopsep=0pt, leftmargin=12pt, itemsep=2pt]
    \item We identify a shared limitation across scalar, multi-attribute, and parametric-distributional RMs: all commit to a fixed output-distribution family. We instead recast reward modeling as density estimation over $p(\mathbf{r}\mid x,y)$.
    \item We introduce \textbf{DRM}, a new reward-modeling paradigm that replaces the value head with a DiT head imposing no parametric form on the output, making it well-suited to the multimodal reward distributions that human preference induces. A single architecture covers both multi-attribute regression and a distributional BT objective on preference pairs, and the head opens a reward-axis test-time scaling absent from current RMs.
    \item We demonstrate the empirical value of DRM across five benchmarks, where it outperforms matched baselines and remains competitive with substantially larger RMs. DRM captures multimodal reward structure associated with human disagreement, while its distributional statistics improve reward-model decisions and downstream RLHF.
\end{itemize}

\section{Method}

\begin{figure*}[t]
    \centering
    \includegraphics[width=\textwidth]{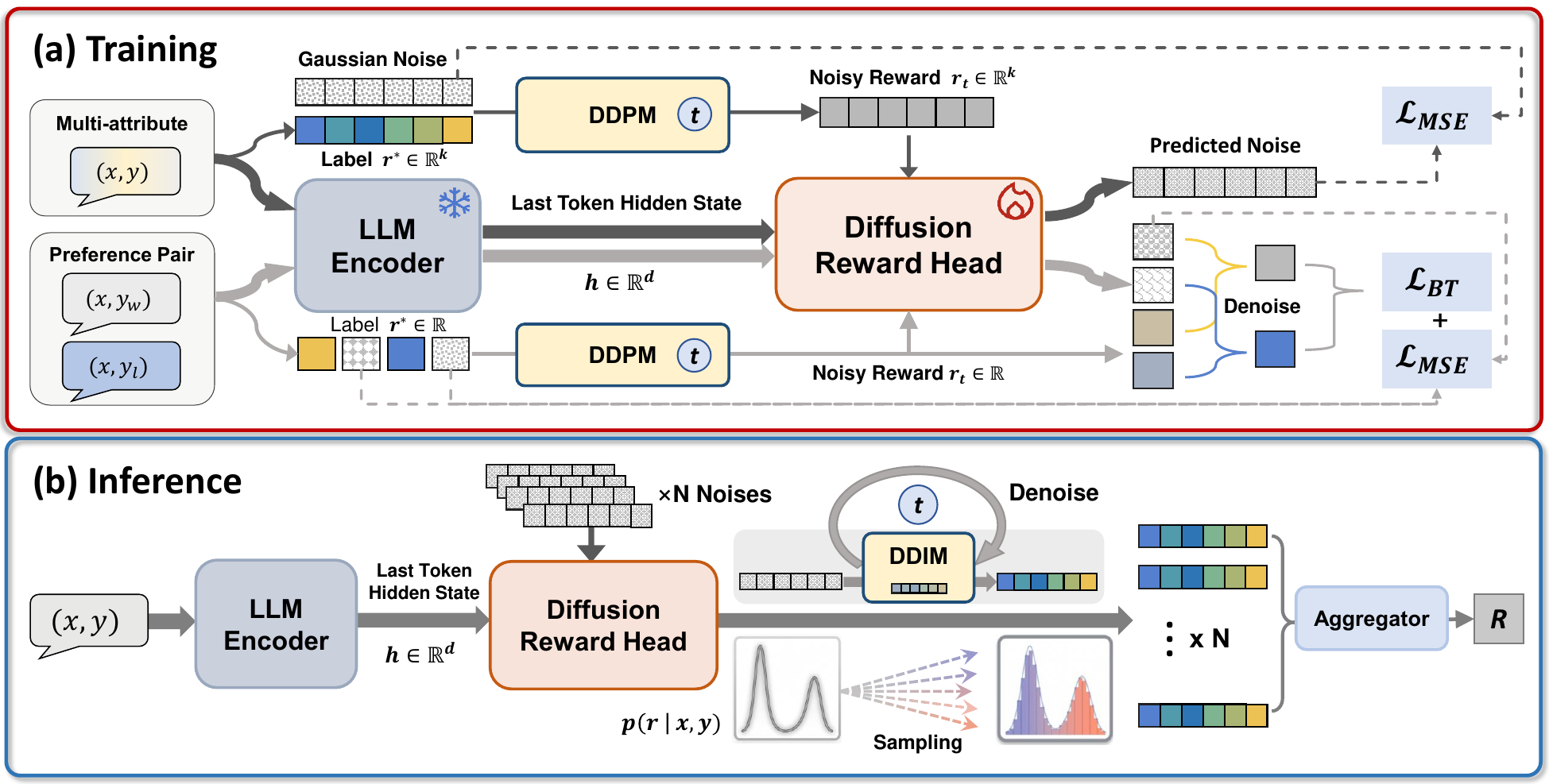}
    \caption{Overview of DRM. A frozen encoder maps $(x,y)$ to a hidden state $\mathbf{h}$ that conditions a Diffusion Reward Head modeling $p(\mathbf{r}\mid x,y)$. \textbf{(a) Training.} A single head supports both multi-attribute regression, supervised by a masked denoising loss $\mathcal{L}_{\mathrm{MSE}}$, and pairwise preference data, supervised by an additional Bradley--Terry objective $\mathcal{L}_{\mathrm{BT}}$ on the denoised rewards. \textbf{(b) Inference.} Drawing $N$ samples via DDIM yields an empirical reward distribution that captures multimodal structure and is summarized by an aggregator into the final reward $R$.}
    \label{fig:overview}
    \vspace{-15pt}
\end{figure*}

DRM replaces the deterministic scalar value head of conventional reward models with a Diffusion Reward Head that directly models $p(\mathbf{r} \mid x, y)$ without committing to any parametric family. As shown in \Cref{fig:overview}, a frozen LLM encoder produces a representation $\mathbf{h}$ of $(x, y)$, a lightweight Diffusion Transformer (DiT) denoises Gaussian noise into reward vectors conditioned on $\mathbf{h}$ (\Cref{sec:arch}), trained either on multi-attribute or pairwise preference data (\Cref{sec:training}), and $N$ samples form an empirical distribution at inference (\Cref{sec:inference}).

\subsection{Architecture}
\label{sec:arch}

We explore a complementary direction to existing reward modeling methods: we model $p(\mathbf{r} \mid x, y)$ implicitly through an iterative denoising process. This is motivated by the well-documented mode coverage of diffusion models~\citep{dhariwal2021diffusion} and their ability to approximate arbitrary continuous densities~\citep{song2020score,lipman2022flow}, which makes them well-suited to the multimodal reward distributions induced by human preference. Concretely, DRM consists of two components: a frozen LLM encoder and a Diffusion Reward Head, followed by a non-parametric aggregator described in \Cref{sec:inference}.

\vspace{+3pt}\vspace{+3pt}\noindent\textbf{LLM Encoder.}
For each prompt--response pair $(x,y)$, we use a frozen language model as the backbone encoder. We format the prompt and response into a single input sequence, feed it into the encoder, and take the hidden state of the last token as the semantic representation of the sample:
\[
    \mathbf{h} = \mathrm{Enc}(x,y) \in \mathbb{R}^{d_{\mathrm{enc}}}.
\]
In our implementation, this encoding process is performed offline: we precompute $\mathbf{h}$ for each $(x,y)$, and then train the Diffusion Reward Head on top of these frozen representations. This decouples costly language representation learning from subsequent reward distribution modeling, preserving the semantic priors of the pretrained encoder while substantially reducing training cost.

\vspace{+3pt}\noindent\textbf{Diffusion Reward Head.}
The DiT reward head $\epsilon_\theta$ is a lightweight DiT that models the conditional distribution of a $K$-dimensional reward vector:
\[
    p_\theta(\mathbf{r} \mid \mathrm{Enc}(x,y)) = p_\theta(\mathbf{r} \mid \mathbf{h}), \qquad \mathbf{r} \in \mathbb{R}^K.
\]
Given a noisy reward $\mathbf{r}_t$ and timestep $t$, the head predicts the noise during the forward process,
\[
    \hat{\boldsymbol{\epsilon}} = \epsilon_\theta(\mathbf{r}_t, t, \mathbf{h}),
\]
where the textual condition $\mathbf{h}$ and timestep $t$ are injected into each DiT block via adaptive layer normalization~\citep{peebles2023scalable}. 
Full implementation details including projection layers, sinusoidal timestep embedding, the conditioning fusion, and initialization are deferred to \Cref{app:dit_details}.

\subsection{Training}
\label{sec:training}

DRM is trained within a unified diffusion framework that supports both multi-attribute regression and pairwise preference data. Both regimes share the same diffusion parameterization and differ only in how supervision targets are constructed.

\vspace{+3pt}\noindent\textbf{Probabilistic Formulation.}
Our goal is to learn the conditional distribution
\[
    p_\theta(\mathbf{r} \mid x, y, \mathbf{m}), \qquad \mathbf{r} \in \mathbb{R}^K,
\]
where the reward space has dimension $K=1$ for scalar reward and $K>1$ for multi-attribute reward vectors, and $\mathbf{m} \in \{0, 1\}^K$ indicates which dimensions are annotated for $(x,y)$. The mask is introduced so that a single model can be trained on multi-attribute datasets within a unified reward space, with unlabeled dimensions excluded from supervision. Since $\mathbf{m}$ is data-specific, we omit it from the notation where it is clear from context.

\vspace{+3pt}\noindent\textbf{Training Objective.}
We define two complementary objectives sharing the same parameterization.

\emph{Multi-attribute regression.} For data with explicit reward annotations $\mathbf{r}_0$, we train the head with a masked denoising loss that minimizes the mean squared error between the predicted and ground-truth noise over the annotated dimensions only:
\[
    \mathcal{L}_{\text{denoise}} = \frac{1}{\|\mathbf{m}\|_1} \bigl\| \mathbf{m} \odot \bigl( \epsilon_\theta(\mathbf{r}_t, t, \mathbf{h}, \mathbf{m}) - \boldsymbol{\epsilon} \bigr) \bigr\|_2^2,
\]
where the forward process is
\(
    \mathbf{r}_t = \sqrt{\bar{\alpha}_t}\, \mathbf{r}_0 + \sqrt{1-\bar{\alpha}_t}\, \boldsymbol{\epsilon},\
    \boldsymbol{\epsilon} \sim \mathcal{N}(\mathbf{0}, \mathbf{I})
\).
Under Gaussian diffusion, this objective is equivalent to learning the score function of the conditional reward distribution~\citep{ho2020denoising,song2020score}.

\emph{Pairwise preference.} For data containing only pairwise preference, pointwise regression is insufficient to capture which response is preferred. For each pair $(x, y_w, y_l)$, let $\hat{r}_0^{(w)}$ and $\hat{r}_0^{(l)}$ denote the denoised reward estimates obtained from the standard reconstruction
\(
    \hat{\mathbf{r}}_0 = (\mathbf{r}_t - \sqrt{1-\bar{\alpha}_t}\, \hat{\boldsymbol{\epsilon}}) / \sqrt{\bar{\alpha}_t}
\).
We impose a Bradley--Terry-style ranking objective on the denoised estimates, and derive its relationship to the corresponding distribution-level preference likelihood in Appendix~\ref{app:loss_design}:
\[
    \mathcal{L}_{\text{BT}} = -\log \sigma\!\left( \hat{r}_0^{(w)} - \hat{r}_0^{(l)} \right).
\]
To preserve the denoising signal alongside the ranking signal, we additionally apply $\mathcal{L}_{\text{denoise}}$ on them. The final pairwise loss is
\[
    \mathcal{L}_{\text{pair}} = \mathcal{L}_{\text{denoise}} + \lambda_{\text{BT}} \mathcal{L}_{\text{BT}}.
\]

Additionally, for preference-only data, absolute reward labels are unavailable. We therefore construct symmetric pseudo-reward targets centered at zero, with a fixed margin \(\Delta\) between the preferred and rejected responses. These targets provide the denoising supervision, while a BT-style auxiliary loss further enforces the pairwise ordering.

\begin{figure}[t]
\begin{minipage}[t]{0.485\textwidth}
\hrule height 0.8pt
\captionof{algorithm}{Training of DRM}
\label{alg:drm_training}
\scriptsize
\vspace{-10pt}
\hrule height 0.6pt
\begin{algorithmic}[1]
\Require Frozen backbone $\mathrm{Enc}$, DiT reward head $\epsilon_\theta$, training set $\mathcal{D}$, noise schedule $\{\bar{\alpha}_t\}_{t=1}^{T}$, CFG rate $p_{\mathrm{drop}}$
\For{each training iteration}
    \State Sample a minibatch $\mathcal{B} \subset \mathcal{D}$
    \For{each sample $(x, y) \in \mathcal{B}$}
        \State Obtain semantic embedding $\mathbf{h} = \mathrm{Enc}(x, y)$
        \State Construct reward target $\mathbf{r}_0$ and mask $\mathbf{m}$
        \State With probability , replace $\mathbf{h}$ with the unconditional vector
        \State Sample $t \sim \mathrm{U}(\{1,\dots,T\})$ and $\boldsymbol{\epsilon} \sim \mathcal{N}(\mathbf{0}, \mathbf{I})$
        \State Form $\mathbf{r}_t = \sqrt{\bar{\alpha}_t}\mathbf{r}_0 + \sqrt{1-\bar{\alpha}_t}\boldsymbol{\epsilon}$
        \State Predict $\hat{\boldsymbol{\epsilon}} = \epsilon_\theta(\mathbf{r}_t, t, \mathbf{h}, \mathbf{m})$
    \EndFor
    \If{$\mathcal{D}$ is multi-attribute}
        \State Compute $\mathcal{L}_{\mathrm{denoise}}$ over the batch
    \Else \Comment{pairwise preference}
        \State Recover $\hat{\mathbf{r}}_0 = (\mathbf{r}_t - \sqrt{1-\bar{\alpha}_t}\hat{\boldsymbol{\epsilon}}) / \sqrt{\bar{\alpha}_t}$ for each response
        \State Compute $\mathcal{L}_{\mathrm{pair}} = \mathcal{L}_{\mathrm{denoise}} + \lambda_{\mathrm{BT}}\mathcal{L}_{\mathrm{BT}}$
    \EndIf
    \State Update $\epsilon_\theta$
\EndFor
\end{algorithmic}
\hrule height 0.8pt
\end{minipage}\hfill
\begin{minipage}[t]{0.485\textwidth}
\hrule height 0.8pt
\captionof{algorithm}{Inference of DRM}
\vspace{-10pt}
\hrule height 0.6pt
\label{alg:drm_inference}
\scriptsize
\begin{algorithmic}[1]
\Require Prompt-response pair $(x,y)$, encoder $\mathrm{Enc}$, DiT reward head $\epsilon_\theta$, DDIM scheduler, sampling steps $S$, number of reward samples $N$, guidance scale $\omega$, aggregation function $\mathcal{A}(\cdot)$
\State Compute the semantic embedding $\mathbf{h} = \mathrm{Enc}(x,y)$
\For{$i=1$ to $N$}
    \State Sample initial Gaussian noise $\mathbf{r}^{(i)}_T \sim \mathcal{N}(\mathbf{0},\mathbf{I})$
    \For{$t$ in DDIM timesteps}
        \State Predict unconditional noise $\hat{\boldsymbol{\epsilon}}_{\mathrm{uncond}} = \epsilon_\theta(\mathbf{r}^{(i)}_t,t,\varnothing)$
        \State Predict conditional noise $\hat{\boldsymbol{\epsilon}}_{\mathrm{cond}} = \epsilon_\theta(\mathbf{r}^{(i)}_t,t,\mathbf{h})$
        \State Apply classifier-free guidance $\hat{\boldsymbol{\epsilon}} = \hat{\boldsymbol{\epsilon}}_{\mathrm{uncond}}  + \omega\left(  \hat{\boldsymbol{\epsilon}}_{\mathrm{cond}} -\hat{\boldsymbol{\epsilon}}_{\mathrm{uncond}}\right)$
        \State Perform one DDIM reverse step $\mathbf{r}^{(i)}_{t-1}  = \mathrm{DDIMStep}(\mathbf{r}^{(i)}_t,\hat{\boldsymbol{\epsilon}},t)$
    \EndFor
    \State Obtain one reward sample $\mathbf{r}^{(i)} \leftarrow \mathbf{r}^{(i)}_0$
\EndFor
\State Form the empirical reward distribution $\hat{P}(\mathbf{r}\mid x,y)$
\State \Return ${Agg}\!\left(\{\mathbf{r}^{(i)}\}_{i=1}^{N}\right)$ \Comment{Aggregation}
\end{algorithmic}
\hrule height 0.8pt
\end{minipage}
\end{figure}

\vspace{+3pt}\noindent\textbf{Training Procedure.}
The full training procedure is summarized in \Cref{alg:drm_training} and visualized in \Cref{fig:overview} (a). During training, we additionally apply classifier-free guidance (CFG)~\citep{ho2022classifier}. With a fixed probability, the textual condition $\mathbf{h}$ is replaced by a learnable unconditional vector, which enables guided sampling at inference time (\Cref{sec:inference}). Training datasets and hyperparameter settings are reported in \Cref{sec:exp_setup}.

\subsection{Inference}
\label{sec:inference}

At inference time, we first estimate the conditional reward distribution by sampling from the trained Diffusion Reward Head, and then summarize it into the statistic required by the downstream task. Unlike conventional reward models that only produce a single scalar, the inference process of DRM naturally preserves distribution-level reward information. Therefore, DRM can be used not only for standard reward scoring, but also for test-time scaling based on distribution.

\vspace{+3pt}\noindent\textbf{Reward Distribution Sampling.}
Given $(x, y)$ and its semantic embedding $\mathbf{h}$, we draw $N$ independent samples from the reward head by running the reverse process from standard Gaussian noise,
\[
    \mathbf{r}^{(1)}, \dots, \mathbf{r}^{(N)} \sim p_\theta(\mathbf{r} \mid \mathbf{h}),
\]
which together form an empirical approximation to the reward distribution. We use DDIM~\citep{song2020denoising} to reduce the number of reverse steps and combine it with CFG controlled by a guidance scale $\omega$; the full procedure is summarized in \Cref{alg:drm_inference} and visualized in \Cref{fig:overview} (b).

\vspace{+3pt}\noindent\textbf{Aggregations.}
After obtaining the empirical distribution, we can exploit this distributional information in ways that distinguish it from conventional scalar reward models.
In this work we use the \textbf{mean} $\bar{\mathbf{r}} = \frac{1}{N}\sum_i \mathbf{r}^{(i)}$ averaged across reward dimensions for an empirical study, which yields a scalar reward compatible with standard RLHF, Best-of-$N$ selection, and reward-benchmark protocols. We study uncertainty-aware rejection and risk-sensitive aggregation in Section~\ref{sec:distribution}, demonstrating that the learned reward distribution provides useful decision signals beyond a single scalar score.

\vspace{+3pt}\noindent\textbf{Two Axes of Test-Time Scaling.}
DRM exposes two independent test-time scaling axes. Along the \emph{response axis}, the classical Best-of-$N$ approach generates $N$ candidate responses and selects the one with the highest aggregated reward, $y^\star = \arg\max_{y_i}\, \bar{r}(x, y_i)$, which is shared with any scalar reward model. Along the \emph{reward axis}, increasing the number of diffusion samples $N$ for a fixed $(x, y)$ tightens the estimate of $\bar{\mathbf{r}}(x, y)$, improving scoring precision without changing the candidate set.\textbf{ The second axis is unique to DRM: a deterministic scalar RM produces the same score for any $N$, so no analogous lever exists.} We empirically study both axes in \Cref{sec:exp_tts}.
\section{Main Evaluation}

\subsection{Experimental Setup}
\label{sec:exp_setup}

We train two DRM variants under the two supervision regimes of \Cref{sec:training}: \textbf{DRM-Multi-8B}, trained on multi-attribute reward data, and \textbf{DRM-Pref-8B}, trained on pairwise preference data. Both DRM variants use the LLM encoder of FsfairX-LLaMA3-RM-v0.1~\citep{dong2023raft} as a frozen backbone, with its scalar value head removed and replaced by our Diffusion Reward Head.

\vspace{+3pt}\noindent\textbf{Training data.}
DRM-Multi-8B is trained on the aggregated multi-attribute reward corpus of ArmoRM~\citep{wang2024interpretable}, which unifies several public reward-modeling sources into $569$K samples annotated over $19$ attributes (helpfulness, coherence, instruction following, etc.), each example labeling only a subset. DRM-Pref-8B is trained on the Tulu3 preference mixture~\citep{lambert2024tulu} ($273$K chosen/rejected pairs, no explicit reward labels, preference margin $\Delta = 1$ is used to get pseudo-rewards). The two regimes follow the masked denoising and distributional Bradley--Terry objectives of \Cref{sec:training} respectively.

\vspace{+3pt}\noindent\textbf{Benchmarks.}
We evaluate on five public reward-model benchmarks: RewardBench v2~\citep{malik2025rewardbench}, PPE~\citep{frick2025evaluate}, RMB~\citep{zhou2025rmb}, RM-Bench~\citep{liu2025rm}, and JudgeBench~\citep{tan2025judgebench}. Together they span chat, instruction following, mathematics, code, reasoning, factuality, and safety, and probe complementary protocols like pairwise accuracy and Best-of-$N$ selection. Per-benchmark statistics and task compositions are deferred to \Cref{app:benchmarks}.

\vspace{+3pt}\noindent\textbf{Baselines.}
We compare against three families of reward models. \textbf{Discriminative RMs} directly output a deterministic scalar score (e.g., ArmoRM-Llama3-8B-v0.1~\cite{wang2024interpretable}, Skywork-Reward-Llama-3.1-8B-v0.2~\cite{Skywork-Reward-Llama-3.1-8B-v0.2} and InternLM2-20B-Reward~\cite{internlm2-7b-reward}). \textbf{Generative RMs} emit a textual verdict or reasoning trace before a score (e.g., DeepSeek-GRM-27B~\cite{DeepSeek-GRM-27B}, the closed-source GPT-4o~\cite{gpt4o} and Claude-3.5-Sonnet~\cite{claude35sonnet}). \textbf{Parametric distributional RMs} model the reward distribution within a fixed family (QRM-Gemma-2-27B~\cite{dorka2024quantile}, URM-Llama-3.1-8B~\cite{lou2024uncertainty}). DRM differs from all three by modeling $p(\mathbf{r}\mid x,y)$ non-parametrically, which additionally yields uncertainty and distributional statistics rather than a point estimate. These comparisons let us ask whether explicit distributional modeling helps on standard reward-modeling tasks, and whether a diffusion head is more expressive than existing parametric-distributional approaches.

\begin{table*}[t]
    \centering
    \scriptsize
    \setlength{\tabcolsep}{4pt}
    \caption{Overall benchmark results of different reward models. Baselines are taken from \citet{liu2025skywork} where available; entries not reported therein are evaluated by us under the same protocol. \textbf{Gray rows denote large-scale or closed baselines that are not directly comparable in model size or publicly reported training-data scale.}}
    \vspace{-8pt}
    \label{tab:main_results}
    \resizebox{\textwidth}{!}{%
    \begin{tabular}{lccccccc}
    \toprule
    \textbf{Reward Models} & \textbf{RewardBench v2} & \textbf{PPE Pref} & \textbf{PPE Corr} & \textbf{RMB Pairwise} & \textbf{RM-Bench} & \textbf{JudgeBench} & \textbf{Avg.} \\
    \midrule
    \multicolumn{8}{l}{\textbf{Discriminative Reward Models}} \\
    \addlinespace[2pt]
    \rowcolor{gray!15} internlm2-7b-reward~\cite{internlm2-7b-reward} & 53.4  & 62.1  & 60.4  & 67.1  & 67.1  & 59.4  & 61.6  \\ 
    Eurus-RM-7b~\cite{Eurus-RM-7b} & 58.1  & 59.6  & 60.0  & 65.5  & 69.0  & 58.4  & 61.8  \\ 
    \rowcolor{gray!15} Starling-RM-34B~\cite{Starling-RM-34B} & 45.5  & 62.8  & 60.3  & 72.0  & 67.1  & 63.8  & 61.9  \\ 
    RM-Mistral-7B~\cite{dong2023raft} & 59.6  & 61.8  & 56.4  & 66.6  & 66.9  & 62.1  & 62.2  \\ 
    ArmoRM-Llama3-8B-v0.1~\cite{wang2024interpretable} & 66.5  & 60.6  & 61.4  & 64.6  & 67.7  & 53.2  & 62.3  \\ 
    \rowcolor{gray!15} internlm2-20b-reward~\cite{internlm2-7b-reward} & 56.3  & 61.0  & 63.0  & 62.9  & 72.1  & 64.3  & 63.3  \\ 
    Llama-3-OffsetBias-RM-8B~\cite{Llama-3-OffsetBias-RM-8B} & 64.8  & 59.2  & {64.1}  & 57.8  & 71.3  & {63.5}  & 63.5  \\ 
    FsfairX-LLaMA3-RM-v0.1~\cite{dong2023raft} & 62.9  & {63.1}  & 61.1  & 70.2  & 71.7  & 56.6  & 64.3  \\ 
    Skywork-Reward-Llama-3.1-8B-v0.2~\cite{Skywork-Reward-Llama-3.1-8B-v0.2} & {71.8}  & 62.2  & 60.7  & 66.6  & 64.7  & 62.9  & 64.8  \\ 
    GRM-Llama3-8B-rewardmodel-ft~\cite{GRM-Llama3-8B-rewardmodel-ft} & 67.7  & 62.1  & 60.0  & 70.2  & 69.9  & 62.3  & 65.4  \\ 
    \rowcolor{gray!15} Llama-3.1-Nemotron-70B-Reward~\citep{wang2024helpsteer2pref}  & 76.7  & 64.2  & 63.2  & 64.9  & 72.2  & 65.8  & 67.8  \\ 
    \rowcolor{gray!15} Skywork-Reward-V2-Qwen3-8B~\citep{liu2025skywork}  & 78.2  & 70.6  & 75.1  & 81.2  & 82.6  & 73.4  & 76.9  \\ 
    \midrule
    
    \multicolumn{8}{l}{\textbf{Generative Reward Models}} \\
    \addlinespace[2pt]
    \rowcolor{gray!15} DeepSeek-GRM-27B~\cite{DeepSeek-GRM-27B} & 64.4  & 64.7  & 59.8  & 69.0  & 72.4  & 63.0  & 65.6  \\ 
    \rowcolor{gray!15} GPT-4o~\cite{gpt4o}  & 64.9  & 67.7  & 67.1  & 73.8  & 73.1  & 59.8  & 67.7  \\ 
    \rowcolor{gray!15} Claude-3.5-Sonnet~\cite{claude35sonnet}  & 64.7  & 67.3  & 69.2  & 70.6  & 74.5  & 64.8  & 68.5  \\ 
    \midrule
    
    \multicolumn{8}{l}{\textbf{Parametric Distributional Reward Models}} \\
    \addlinespace[2pt]
    \rowcolor{gray!15} QRM-Gemma-2-27B~\cite{dorka2024quantile} & 76.7  & 52.3  & 54.8  & 53.4  & 65.9  & 57.5  & 60.1  \\ 
    QRM-Llama3.1-8B-v2~\cite{dorka2024quantile} & 70.7  & 57.2  & 60.3  & 61.1  & {72.5}  & 62.6  & 64.1  \\ 
    URM-LLaMa-3.1-8B~\cite{lou2024uncertainty} & {73.9}  & 60.2  & 60.4  & 65.7  & {72.0}  & {64.1}  & {66.1}  \\ 
    \rowcolor{gray!15} LDL-Reward-Gemma-2-27B-v0.1~\cite{LDL-Reward-Gemma-2-27B-v0.1} & 72.5  & 62.4  & 63.9  & 67.9  & 71.0  & 64.2  & 67.0  \\ 
    \midrule
    
    \multicolumn{8}{l}{\textbf{Diffusion Reward Models (Ours)}}\\
    \addlinespace[2pt]
    \textbf{DRM-Multi-8B} & 65.6 & 62.5 & {63.8 }& {78.0} & 68.8 & 58.6 & {66.2} \\
    \textbf{DRM-Pref-8B} & 65.7 & {63.0} & 62.5 & {78.2} & 68.1 & 57.1 & 65.8 \\
    \bottomrule
    \end{tabular}
    }
    \vspace{-10pt}
\end{table*}

\begin{wraptable}{r}{0.50\textwidth}
\vspace{-10pt}
\centering
\small
\caption{Configurations of the two DRM variants.}
\resizebox{0.50\textwidth}{!}{%
\begin{tabular}{lcc}
\toprule
 & \textbf{DRM-Multi-8B} & \textbf{DRM-Pref-8B} \\
\midrule
Training Data        & ArmoRM agg.        & Tulu3 pref. \\
Reward Dim $K$        & 19                 & 1 \\
\addlinespace[2pt]
\multicolumn{3}{l}{\textbf{Diffusion Reward Head}} \\
\addlinespace[2pt]
Hidden Size          & 384                & 384 \\
Blocks / Heads       & 3 / 6              & 3 / 6 \\
Dropout              & 0.2                & 0.2 \\
\addlinespace[2pt]
\multicolumn{3}{l}{\textbf{Training}} \\
\addlinespace[2pt]
Max Diffusion Steps      & 1000               & 1000 \\
Beta Schedule        & sqcos & sqcos \\
Batch Size           & 64                 & 128 \\
Learning Rate        & $5{\times}10^{-5}$ & $3{\times}10^{-5}$ \\
$\lambda_{\mathrm{BT}}$ & --              & 0.5 \\
\addlinespace[2pt]
\multicolumn{3}{l}{\textbf{Inference}} \\
\addlinespace[2pt]
DDIM Steps           & 10                 & 10 \\
Guidance $\omega$    & 7                  & 7 \\
\# samples $N$       & 32                 & 32 \\
\bottomrule
\end{tabular}
}
\vspace{-20pt}
\label{tab:drm_config}
\vspace{-20pt}
\end{wraptable}

\vspace{+3pt}\noindent\textbf{DRM configuration.}
Both variants share the same architecture and diffusion schedule, differing only in reward dimension $K$ and a few optimization settings. Full configurations are listed in \Cref{tab:drm_config}. All hyperparameters are  fixed across benchmarks. The hyperparameter search is detailed in \Cref{app:hparam}.

\subsection{Main Results}
\label{sec:exp_main}

\vspace{+1pt}\noindent\textbf{Under matched data and backbone, the diffusion head wins.}
\Cref{tab:main_results} reports results across the five benchmarks. We position DRM as a first exploration of a native distributional reward head, so the most controlled comparison is against models trained on the same ArmoRM corpus with the same backbone, where only the reward head differs. Here DRM-Multi-8B reaches an average of $66.2$, exceeding both the scalar multi-attribute head ArmoRM ($62.3$) and the parametric-quantile head QRM ($64.1$); the gain is largest on RMB ($78.0$). With data and backbone held fixed, this isolates the reward head as the source of the improvement.

\vspace{+1pt}\noindent\textbf{DRM stays competitive beyond the matched setting.}
Beyond this controlled comparison, DRM remains competitive with strong discriminative RMs trained on other data or larger backbones, such as Eurus-RM-7B ($61.8$), Skywork-Reward-Llama-3.1-8B-v0.2 ($64.8$), and Llama-3.1-Nemotron-70B ($67.8$), and approaches or even surpasses generative judges that rely on far larger capacity and explicit reasoning at much higher cost, e.g.\ DeepSeek-GRM-27B ($65.6$) and GPT-4o ($67.7$). Despite its modest training scale, a non-parametric diffusion head is thus competitive across discriminative, distributional, and generative families.

\vspace{+1pt}\noindent\textbf{A non-parametric head matches parametric ones while staying more general.}
The parametric distributional models share DRM's goal of modeling $p(\mathbf{r}\mid x,y)$, but realize it with an MLP that emits the parameters of a fixed family in a single forward pass, like a Gaussian for URM or a fixed quantile grid for QRM. DRM instead represents the distribution implicitly through iterative denoising, with no parametric form imposed on its shape. It performs on par with these models on average ($66.2$, comparable to URM's $66.1$ and QRM's $64.1$) while making no distributional assumption, suggesting diffusion reward modeling offers a flexible and viable alternative.

\vspace{+1pt}\noindent\textbf{The framework transfers across supervision types.}
Finally, DRM-Multi-8B and DRM-Pref-8B perform similarly ($66.2$ vs.\ $65.8$) despite being trained under multi-attribute regression versus pairwise preference, indicating that the same reward-space diffusion framework transfers across both regimes. We analyze this gap in Appendix \ref{sec:training_scale}, where size-matched experiments highlight the effect of training scale.

\subsection{Fine-Grained Analysis of DRM}

\vspace{+3pt}\noindent\textbf{Balanced preference.} 
In RMB Best-of-$N$ evaluation (\Cref{tab:RMB}), DRM maintains high and closely matched scores on both Helpfulness and Harmlessness, with an average difference of less than 5 points. In contrast to some reward models that often score high on one dimension but noticeably lower on the other, DRM’s balanced performance indicates that its reward representations can encode multimodal preferences simultaneously. Moreover, the two DRM variants show highly consistent performance across both dimensions, further suggesting that this balance reflects the intrinsic multimodal preference modeling capability of the method rather than a coincidental outcome.

\begin{wraptable}{r}{0.5\textwidth}
    \vspace{-10pt}
    \centering
    \caption{
    Downstream RLHF performance with different reward models. All methods use the same Tulu3-8B-SFT initialization and RLHF training setup.
    }
    \resizebox{0.5\textwidth}{!}{%
    \begin{tabular}{lcc}
    \toprule
        Actor & Arena-Hard v2 & MT-Bench \\ \midrule
        Tulu3-8B-SFT & 1.0 & 59.4 \\ 
        + ArmoRM RLHF & 1.0 & 71.5 \\
        + FsfairX RLHF & 1.3 & 73.6 \\ 
        + DRM-Multi RLHF & 2.0 & 74.8 \\ \bottomrule
    \end{tabular}
    }
    \label{tab:rlhf}
    \vspace{-10pt}
\end{wraptable}

\vspace{+3pt}\noindent\textbf{Correctness preference judgment.}
We use JudgeBench (\Cref{tab:JudgeBench}) and PPE Correctness (\Cref{tab:PPE_Corr}) to analyze whether DRM can capture objective correctness. DRM-Multi-8B achieves an overall score of 58.6 on JudgeBench, outperforming GPT-4o and Claude-3.5-Sonnet. On PPE Correctness, DRM-Multi-8B also reaches 63.8, surpassing multiple open-source reward models and maintaining stable performance on subtasks such as math, MMLU, and MBPP. These results indicate that the reward representations learned by DRM do not merely reflect response style, politeness, or general helpfulness, but can also identify correctness differences in factual, knowledge, mathematical, and code to some extent. This is important for reward models, since real world preferences often involve both subjective quality and objective correctness.

\subsection{DRM as a Training-Time Reward for RLHF}

To further evaluate the effectiveness of DRM, we conduct an RLHF experiment. Specifically, we use allenai/Llama-3.1-Tulu-3-8B-SFT as the common initial actor and perform RLHF training on prompts from UltraFeedback. More details can be seen in Appendix~\ref{sec:rlhf_details}. We compare FsfairX, ArmoRM and DRM-Multi under the same actor initialization and RLHF training setup, and evaluate the resulting actors on Arena-Hard v2 and MT-Bench. The results are shown in Table~\ref{tab:rlhf}. Compared with the scalar FsfairX and ArmoRM reward baseline, DRM-Multi achieves better downstream performance on both benchmarks. On Arena-Hard v2, DRM-Multi-RLHF improves the score from 1.3 with FsfairX-RLHF to 2.0; on MT-Bench, it improves the score from 73.6 to 74.8. These results further demonstrate that the advantages of DRM in reward modeling extend beyond offline reward evaluation and can translate into tangible gains in downstream policy optimization.

\section{Distributional Analysis of DRM}

\subsection{Distributional Validation with Human Disagreement}

To directly validate whether DRM captures distributions of human disagreement, we use datasets with repeated human annotations to evaluate DRM’s distribution modeling ability.

\vspace{+3pt}\noindent\textbf{Structured disagreement in human annotations.} 
We analyze 2 datasets. HelpSteer2-Disagreements provides multiple pointwise ratings for the same prompt-response, while MultiPref provides multiple pairwise preference judgments for the same response pair. The former allows us to examine empirical score distributions for individual responses, while the latter captures annotator disagreement from the pairwise preference perspective. Detailed results are in Table~\ref{tab:disagreement}.

On HelpSteer2-Disagreements, we treat the 0-4 ratings for each response as an empirical distribution and measure rating range, separated clusters, and low-high polarization. For helpfulness, 43.49\% of examples have a rating range of at least 2, 28.20\% contain separated clusters, and 24.34\% contain both low and high ratings. The corresponding numbers for correctness are 41.99\%, 27.79\%, and 24.24\%. These results show that annotator disagreement often exhibits clear clustered or polarized structure rather than only small fluctuations around a mean.

MultiPref shows a similar pattern. We map its 5-level pairwise preferences to signed margins in ${-2,-1,0,1,2}$. For overall preference, 45.64\% of examples contain opposite-side judgments and 37.23\% show separated clusters; for helpfulness, the corresponding ratios are 42.77\% and 34.37\%. This pattern is also strongly attribute-dependent: 87.11\% of harmlessness examples receive fully consistent annotations. Thus, the observed clustering and polarization are not artifacts of the annotation format, but are more pronounced on subjective attributes. These repeated-annotation results provide direct data-side motivation for DRM: human feedback on the same input often exhibits clear instance-level disagreement structure.

\begin{table}[!t]
    \vspace{-10pt}
    \centering
    \caption{Structured disagreement in repeated human annotations on different datasets.}
    \resizebox{\textwidth}{!}{
    \begin{tabular}{lcccc}
    \toprule
        \textbf{Dataset} & \textbf{Dimension} & \textbf{Key Disagreement Statistic(\%)} & \textbf{Separated Clusters(\%)} & \textbf{Polarization(\%)} \\ \midrule
        \textbf{HelpSteer2} & Helpfulness & Range$\geq$2: 43.49 & 28.20 & 24.34 \\ 
        \textbf{HelpSteer2} & Correctness & Range$\geq$2: 41.99 & 27.79 & 24.24 \\ 
        \textbf{MultiPref} & Overall & Opposite-side: 45.64 & 37.23 & 11.40 \\ 
        \textbf{MultiPref} & Helpfulness & Opposite-side: 42.77 & 34.37 & 9.94 \\ \bottomrule
    \end{tabular}
    }
    \label{tab:disagreement}
    \vspace{-10pt}
\end{table}

\vspace{+3pt}\noindent\textbf{DRM aligns with empirical human reward distributions.}
We evaluate DRM-Multi on HelpSteer2-Disagreements. For each prompt-response pair, we sample rewards from DRM and compare them with the empirical distribution formed by repeated human ratings. We measure distributional distance using Wasserstein distance, Jensen--Shannon divergence, and $L_1$ distance, and compare against three simple baselines: an empirical global prior, a global Gaussian, and a pointwise baseline. Results are shown in Table~\ref{tab:distribution_distance}.

On helpfulness, DRM achieves the lowest distance on all three metrics. Its Wasserstein distance is 0.804, compared with 1.030 for the empirical prior, 1.032 for the global Gaussian, and 1.032 for the pointwise baseline; its JS divergence and $L_1$ distance are 0.215 and 0.907, respectively. This shows that DRM does not simply learn a global uncertainty pattern, but adapts its output distribution to each prompt-response pair. On correctness, DRM again achieves the best Wasserstein distance at 0.846 and substantially outperforms the pointwise baseline. Overall, these results show that DRM learns input-conditional reward distributions that better capture the structure of human rating distributions than point predictions or simple global distributions.

\begin{table}[!t]
    \centering
    \caption{Distributional distances to empirical human reward distributions on HelpSteer2-Disagreements. Lower is better for all metrics, and the best results are shown in bold.}
    \begin{tabular}{lcccc}
    \toprule
        \textbf{Model} & \textbf{Dimension} & \textbf{Wasserstein$\downarrow$} & \textbf{JS$\downarrow$} & \textbf{L1$\downarrow$} \\ \midrule
        \textbf{DRM} & Helpfulness & \textbf{0.804} & \textbf{0.215} & \textbf{0.907} \\ 
        \textbf{Empirical prior} & Helpfulness & 1.030 & 0.231 & 1.012 \\ 
        \textbf{Global Gaussian} & Helpfulness & 1.032 & 0.254 & 1.054 \\ 
        \textbf{Pointwise} & Helpfulness & 1.032 & 0.254 & 1.407 \\ \midrule
        \textbf{DRM} & Correctness & \textbf{0.846} & 0.251 & 1.006 \\ 
        \textbf{Empirical prior} & Correctness & 1.019 & \textbf{0.224} & \textbf{0.994} \\ 
        \textbf{Global Gaussian} & Correctness & 1.024 & 0.252 & 1.052 \\ 
        \textbf{Pointwise} & Correctness & 0.995 & 0.415 & 1.386 \\ \bottomrule
    \end{tabular}
    \label{tab:distribution_distance}
    \vspace{-10pt}
\end{table}

\vspace{+3pt}\noindent\textbf{Multimodality tracks human disagreement.}
We further examine whether DRM becomes more multimodal as human disagreement increases. We discretize DRM samples into the same 0-4 rating space as HelpSteer2-Disagreements and classify an output as multimodal when its samples occupy multiple separated rating regions. We then group examples by the level of disagreement in repeated human ratings.

\begin{wrapfigure}{r}{0.45\textwidth}
    \vspace{-10pt}
    \centering
    \includegraphics[width=\linewidth]{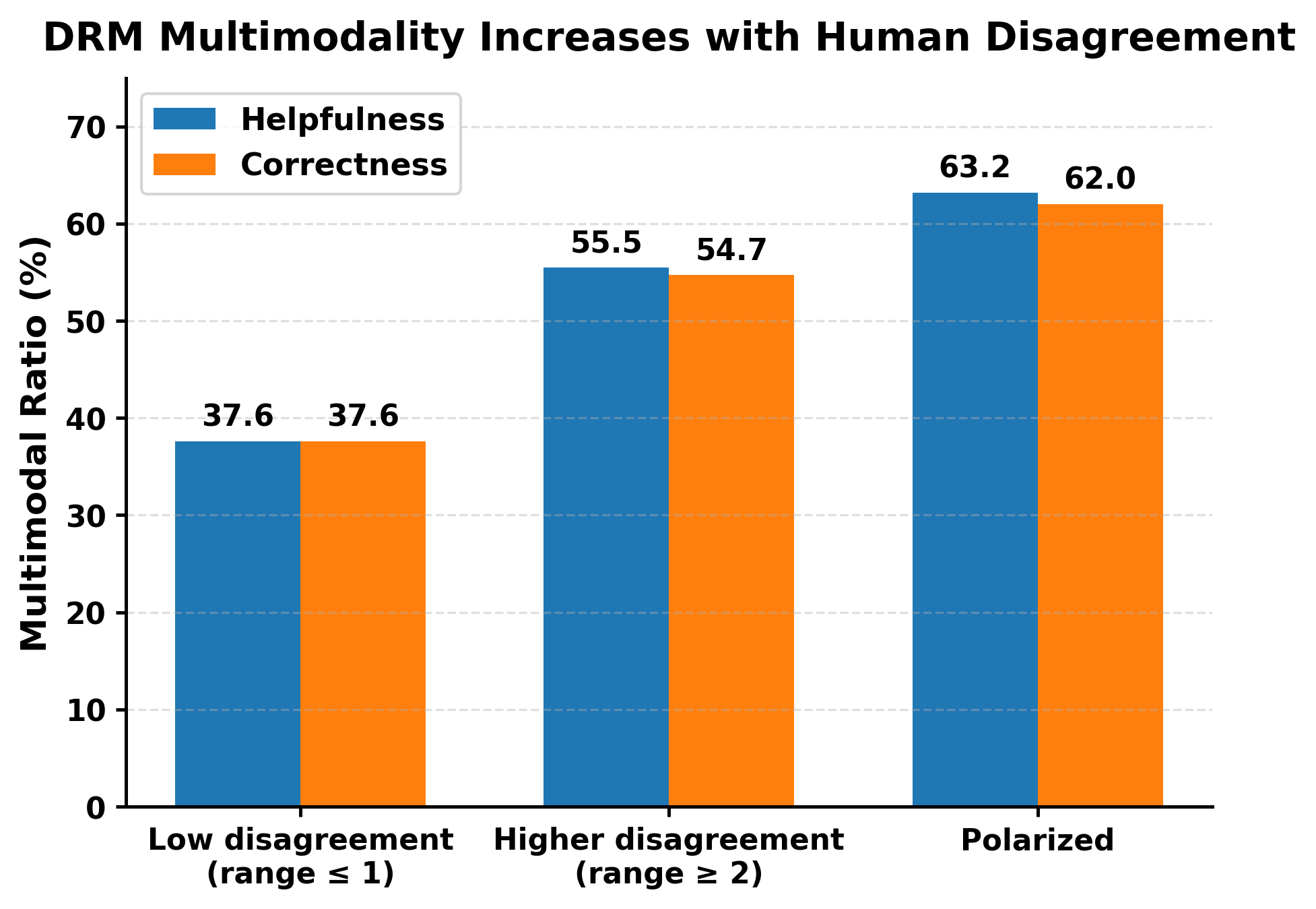}
    \vspace{-20pt}
    \caption{
    DRM multimodal ratio increases with human disagreement.
    }
    \vspace{-20pt}
    \label{fig:disagreement}
\end{wrapfigure}

A clear trend emerges as shown in Figure~\ref{fig:disagreement}. For helpfulness, the multimodal ratio is 37.6\% when the human rating range is at most 1, increases to 55.5\% when the range is at least 2, and further rises to 63.2\% for low-high polarized examples. Correctness shows a similar pattern: 37.6\% $\rightarrow$ 54.7\% $\rightarrow$ 62.0\%. These results show that DRM's multimodal outputs occur more frequently on examples with stronger human disagreement and polarization, indicating that its learned distributions capture distributional signals associated with human disagreement.

Overall, these results consistently show that DRM captures structured reward distributions associated with human disagreement. Repeated human annotations frequently exhibit separated and polarized patterns, while DRM reflects these instance-level structures and becomes increasingly multimodal as human disagreement grows.

\subsection{Distribution-Aware Decision Making}
\label{sec:distribution}

Beyond providing a richer description of human feedback, we further ask whether DRM’s learned distributions can directly improve reward-model decisions. We therefore construct distribution-aware decision rules from uncertainty and distributional statistics, and evaluate them in two settings: when abstention is allowed, we use distributional information to identify unreliable decisions; when all candidates must be ranked, we use a lower-confidence bound for risk-sensitive aggregation.

\vspace{+3pt}\noindent\textbf{Uncertainty-aware rejection.}
We first examine whether DRM’s learned reward distributions can indicate the reliability of reward-model decisions. For Best-of-N, we rank candidates by mean reward, select the top candidate and runner-up, and estimate the probability that the runner-up exceeds the current winner using their reward samples with $ U_{\mathrm{BoN}} = \hat P(r_{\mathrm{runner}} > r_{\mathrm{top}}). $ A larger \(U_{\mathrm{BoN}}\) indicates that the mean-selected winner is less stable under the full reward distributions. For pairwise evaluation, we first compute $ p=\hat P(r_{\mathrm{chosen}}>r_{\mathrm{rejected}}), $ and define a symmetric uncertainty score $ U_{\mathrm{pair}} = 1-|2p-1|, $ where a larger \(U\) indicates that the two reward distributions are harder to distinguish.

We then rank examples by the distributional uncertainty above and reject the most uncertain decisions first. The results are shown in Table~\ref{tab:uncertainty_reject}. As coverage decreases from 100\% to 70\%, accuracy on the retained examples improves substantially: PPE Correctness gains 2.81 percentage points on average across five subtasks, while RMB improves by 4.56-7.31 points across Helpfulness/Harmlessness BoN and pairwise settings. These results show that DRM's full reward distributions provide decision-reliability information beyond mean reward, enabling selective prediction to identify unstable decisions and improve automatic decision accuracy.

\begin{table}[!t]
    \vspace{-5pt}
    \centering
    \caption{
    Selective prediction on PPE Correctness using DRM's distributional uncertainty.
    Performance is reported at different coverage levels, with \textit{Gain} denoting the improvement from 100\% to 70\% .
    }
    \vspace{-5pt}
    \begin{tabular}{lccccc}
    \toprule
        Task & 100\% & 90\% & 80\% & 70\% & Gain \\ \midrule
        GPQA (Best@k) & 44.73 & 45.34 & 46.83 & 48.32 & +3.59 \\ 
        MATH (Best@k) & 50.78 & 53.15 & 55.37 & 56.98 & +6.20 \\ 
        MMLU-Pro (Best@k) & 60.94 & 62.26 & 63.66 & 62.85 & +1.91 \\ 
        IFEval (Best@k) & 62.89 & 62.69 & 62.93 & 64.25 & +1.36 \\ 
        MBPP+ (Best@k) & 69.43 & 69.96 & 70.20 & 70.42 & +0.99 \\ 
        Average & - & - & - & - & +2.81 \\ \bottomrule
    \end{tabular}
    \label{tab:uncertainty_reject}
    \vspace{-5pt}
\end{table}

\begin{table}[!t]
    \centering
    \caption{
    Selective prediction results on RMB at different coverage levels.
    \textit{Gain} denotes the improvement from 100\% to 70\% coverage.
    }
    \vspace{-5pt}
    \begin{tabular}{lccccc}
    \toprule
        Setting & 100\% & 90\% & 80\% & 70\% & Gain \\ \midrule
        Helpfulness BoN & 65.67 & 67.91 & 69.77 & 72.15 & +6.48 \\ 
        Harmlessness BoN & 61.24 & 64.41 & 66.54 & 68.23 & +6.99 \\ 
        Helpfulness Pairwise & 80.80 & 83.58 & 86.02 & 88.11 & +7.31 \\ 
        Harmlessness Pairwise & 73.97 & 76.33 & 77.61 & 78.52 & +4.56 \\ \bottomrule
    \end{tabular}
    \vspace{-10pt}
\end{table}

\vspace{+3pt}\noindent\textbf{Distribution-aware full-coverage ranking.}
In many Best-of-N settings, the system must return a final choice for every example and cannot abstain from high-uncertainty decisions. We therefore study whether DRM’s full reward distribution can improve candidate ranking while maintaining 100\% coverage. Specifically, we use a lower-confidence-bound (LCB) score that jointly considers mean reward and distributional dispersion: $ \mathrm{LCB}_{\lambda} = \mu - \lambda\sigma , $ where \(\mu\) and \(\sigma\) are the mean and standard deviation of DRM reward samples, and we set \(\lambda=0.4\). Compared with mean-only ranking, LCB penalizes candidates with high average reward but unstable reward distributions, thereby accounting for both reward level and distributional stability. The results are shown in Table~\ref{tab:lcb_result}.

The results show that LCB consistently outperforms mean aggregation across all evaluated PPE and RMB Best-of-N settings. On PPE BoN, LCB achieves positive gains for candidate set sizes \(N=2,4,8,16,32\); at \(N=32\), the macro score improves from 57.637 to 57.797. On RMB, LCB also outperforms the mean for both Helpfulness and Harmlessness at \(K=2,3\). For example, at \(K=3\), Helpfulness improves from 68.231 to 68.611, while Harmlessness improves from 62.493 to 62.791.

\subsection{Test-Time Scaling: Two Axes}
\label{sec:exp_tts}

\begin{table}[!t]
    \vspace{-5pt}
    \centering
    \caption{
    Risk-sensitive Best-of-N selection using lower confidence bound (LCB) aggregation.
    LCB consistently improves over mean aggregation across PPE and RMB settings.
    }
    \begin{tabular}{lcccc}
    \toprule
        Setting & Candidates number & Mean & LCB-0.40 & Gain \\ \midrule
        PPE BoN Macro & 2 & 51.211 & 51.274 & +0.063 \\ 
        PPE BoN Macro & 4 & 54.328 & 54.445 & +0.117 \\ 
        PPE BoN Macro & 8 & 56.236 & 56.401 & +0.165 \\ 
        PPE BoN Macro & 16 & 57.381 & 57.556 & +0.175 \\ 
        PPE BoN Macro & 32 & 57.637 & 57.797 & +0.160 \\ \midrule
        RMB Harmlessness BoN & 2 & 74.478 & 74.717 & +0.239 \\ 
        RMB Harmlessness BoN & 3 & 62.493 & 62.791 & +0.298 \\ 
        RMB Helpfulness BoN & 2 & 79.991 & 80.275 & +0.284 \\ 
        RMB Helpfulness BoN & 3 & 68.231 & 68.611 & +0.380 \\ \bottomrule
    \end{tabular}
    \label{tab:lcb_result}
    \vspace{-10pt}
\end{table}

\begin{figure*}[t]
    \centering
    \includegraphics[width=\textwidth]{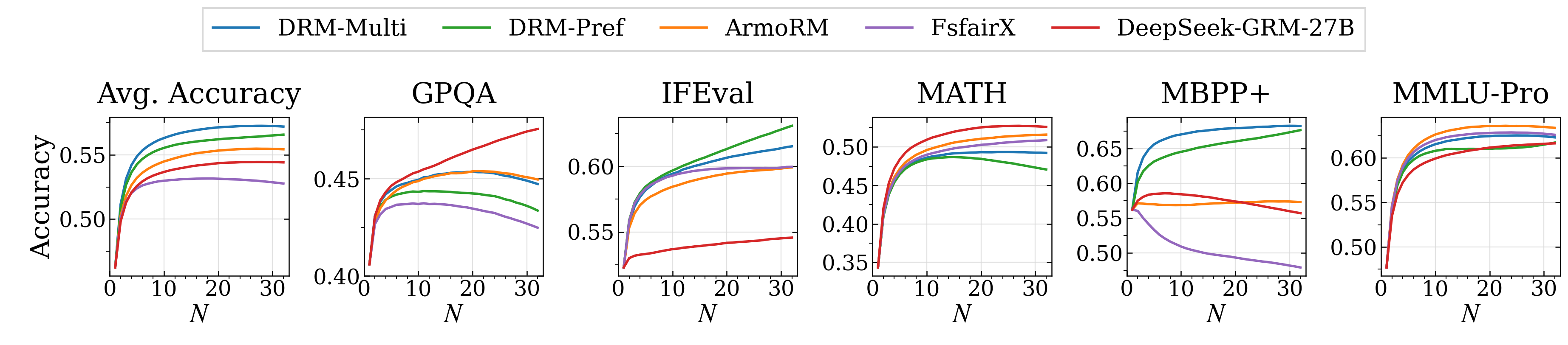}
    \vspace{-30pt}
    \caption{
    Best-of-$N$ scaling curves on PPE Correctness, shown for the five-task average and each individual task.
    }
    \vspace{-10pt}
    \label{fig:BoN_PPE}
\end{figure*}

\begin{wrapfigure}{r}{0.45\textwidth}
    \vspace{-12pt}
    \centering
    \includegraphics[width=\linewidth]{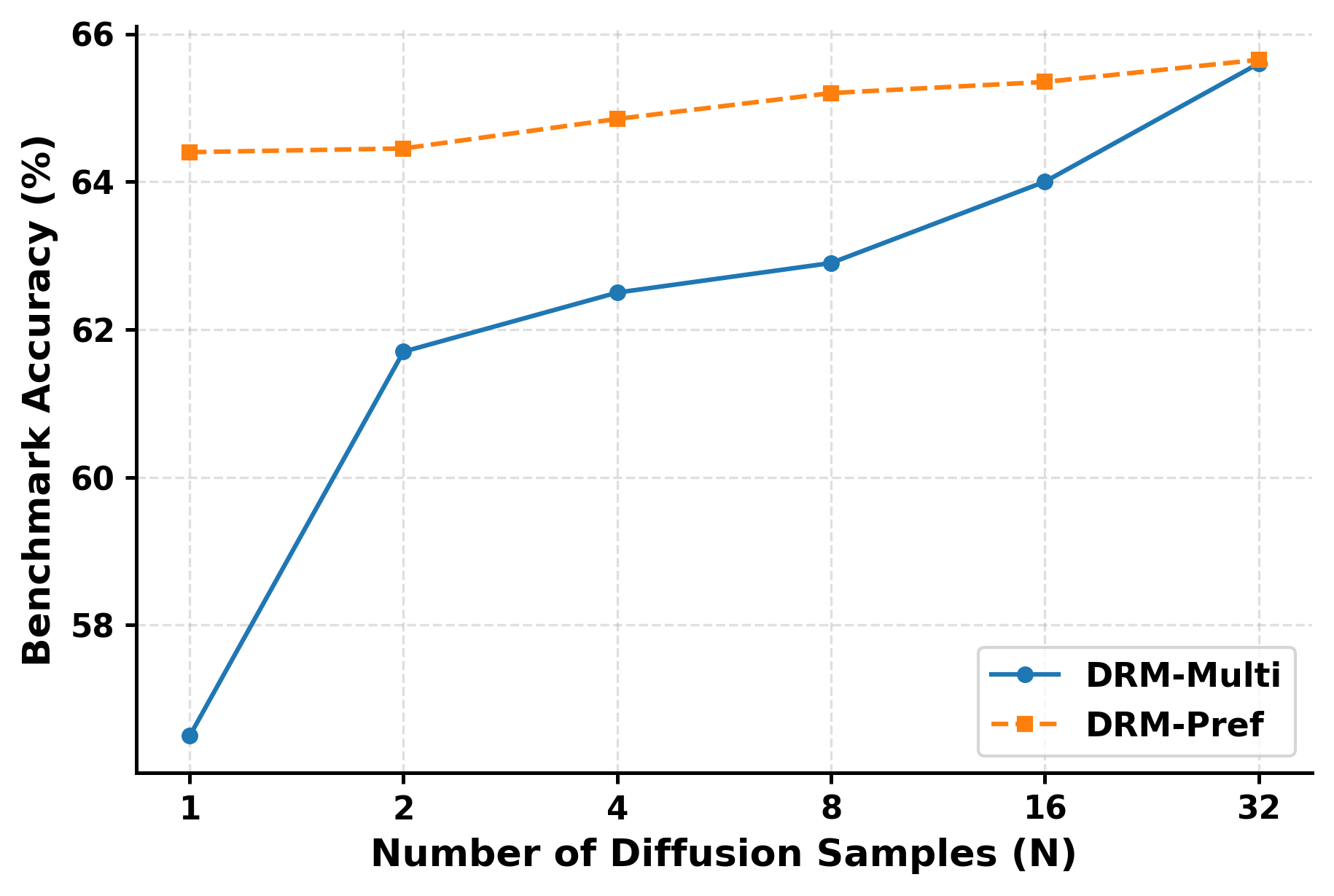}
    \vspace{-20pt}
    \caption{
    Reward-axis scaling on RewardBench v2.
    }
    \vspace{-20pt}
    \label{fig:tts-b}
\end{wrapfigure}

These results show that even at full coverage, the dispersion of DRM’s reward distribution provides useful ranking information beyond the mean. By directly incorporating distributional statistics into reward aggregation, DRM further improves Best-of-N candidate selection, demonstrating that learning the full reward distribution not only captures human disagreement but also directly improves downstream reward-model decisions.

\vspace{+3pt}\noindent\textbf{Response-axis scaling: Best-of-$N$ selection.}
Along the response axis, the RM scores $N$ candidate responses and selects the highest, a lever shared by any scalar RM. We evaluate two DRM variants with ArmoRM, FsfairX, and DeepSeek-GRM-27B on PPE Correctness. \Cref{fig:BoN_PPE} shows BoN curves on five challenging PPE Correctness tasks together with their average. Averaged over the five tasks, both DRM variants scale best among all evaluated models, and their accuracy improves monotonically as $N$ grows. 
In contrast, some baselines plateau or even drop at large $N$, where the highest score may pick the wrong answer, which is a reward-hacking trend that DRM may avoid.

\vspace{+3pt}\noindent\textbf{Reward-axis scaling: sampling the reward distribution.}
The reward axis is unique to DRM: for a fixed $(x,y)$, drawing more samples from the diffusion head tightens the estimate of its distribution. As shown in \Cref{fig:tts-b}, DRM-Multi-8B rises from $56.5\%$ at a single sample to $65.6\%$ at $32$ samples on RewardBench v2, while DRM-Pref-8B improves from $64.4\%$ to $65.7\%$. The number of diffusion samples thus acts as a test-time knob that trades computation for scoring precision without retraining, a lever no scalar or single-pass parametric RM provides.

\subsection{Qualitative Illustration of Multimodal Rewards}
\label{sec:exp_multimodal}

\begin{wrapfigure}{r}{0.45\textwidth}
    \vspace{-20pt}
    \centering
    \includegraphics[width=\linewidth]{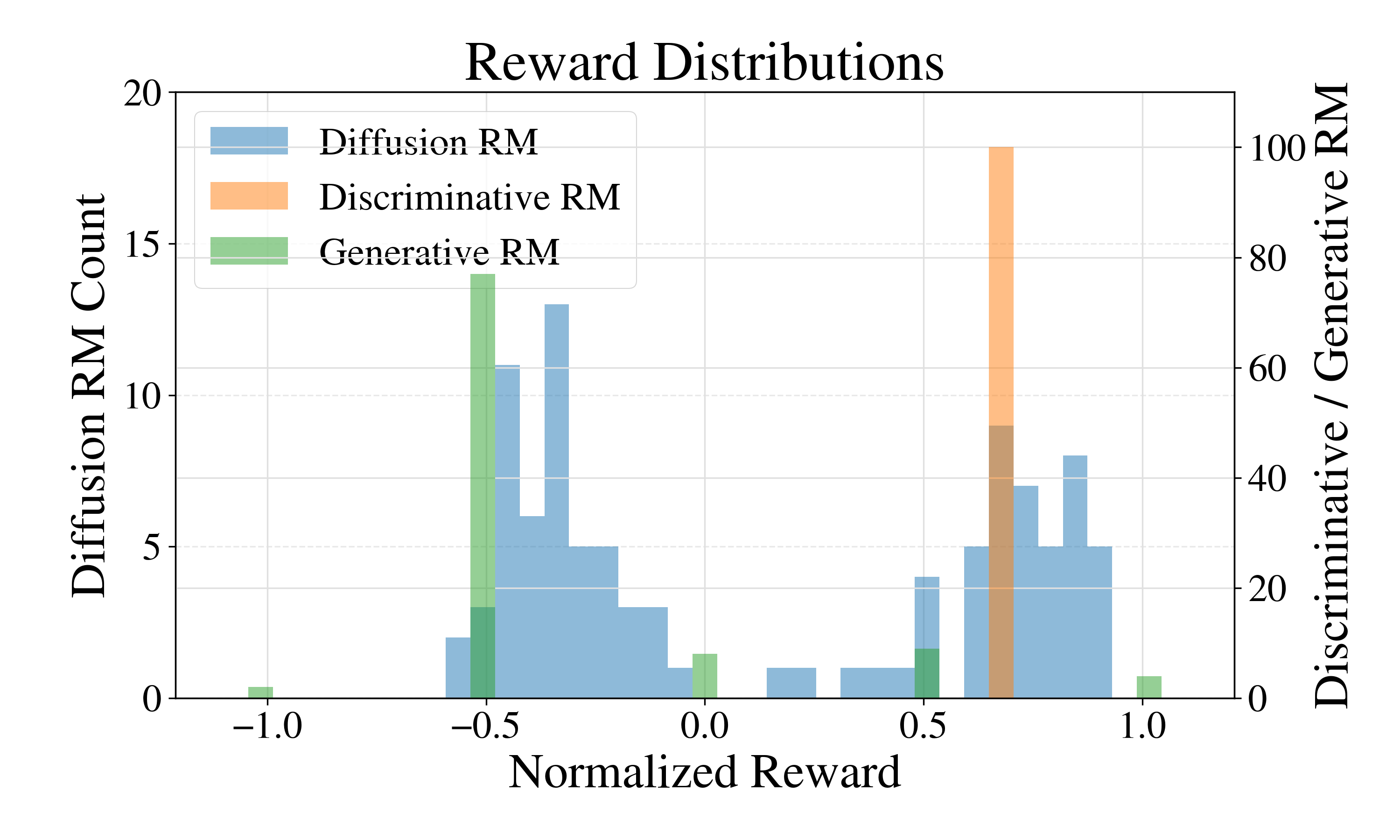}
    \vspace{-20pt}
    \caption{Reward distributions of diffusion, discriminative, and generative reward models.}
    \label{fig:reward_distribution}
    \vspace{-10pt}
\end{wrapfigure}

In this section, we illustrate our central claim that human preference is multimodal and that a non-parametric head can represent what a scalar or unimodal head cannot, on a specific prompt--response pair. We collect $100$ scores each from DRM, a discriminative RM (ArmoRM), and a generative RM (DeepSeek-GRM-27B), compared on a common reward dimension and normalized to $[-1,1]$. The three expose uncertainty differently: DRM's $100$ scores are independent draws from the distribution its diffusion head models; ArmoRM is deterministic and collapses to a single point; DeepSeek-GRM varies only through decoding temperature, i.e.\ noise in generating a verdict rather than a modeled reward distribution.

As shown in \Cref{fig:reward_distribution}, DRM places mass on several well-separated reward regions, recovering a clearly multimodal distribution, while ArmoRM concentrates at a single point and DeepSeek-GRM remains sharply unimodal. This is exactly what the framework predicts: heads that emit a scalar, or that derive randomness only from decoding, commit to one dominant judgment per pair, whereas DRM retains the multiple defensible verdicts a contested example admits. As a single-example illustration this evidence is qualitative, but it shows directly the multimodal reward structure that motivates DRM and that conventional heads cannot express.
\section{Related Work}

\paragraph{Reinforcement Learning from Human Feedback.}
RLHF~\citep{christiano2017deep,stiennon2020learning,ouyang2022training,bai2022training} is the standard recipe for post-training LLMs, spanning PPO~\citep{schulman2017proximal}, GRPO~\citep{shao2024deepseekmath}, reward-ranked fine-tuning~\citep{dong2023raft}, and direct preference methods such as DPO~\citep{rafailov2023direct}, all relying on a learned reward to fit human feedback. RLVR-style training with verifiable rewards~\citep{lambert2024tulu,guo2025deepseek} is an exception in domains with ground-truth answers, but in general-domain alignment reward quality bounds alignment quality~\citep{gao2023scaling,casper2023open}. Rather than proposing a new RL algorithm, we ask what form the learned reward should take.

\paragraph{Multimodal Structure of Human Preference.}
A growing body of work documents that human preference is multimodal instead of a single point. Inter-annotator agreement on RLHF datasets is only $63$--$73\%$ across Anthropic-HH~\citep{bai2022training}, OpenAI summarization~\citep{stiennon2020learning}, and InstructGPT~\citep{ouyang2022training}, with most disagreement reflecting genuine individual preferences rather than noise~\citep{zhang2024diverging}; preference also varies across cultures and demographics~\citep{kirk2024prism}, and alignment further reduces this diversity~\citep{sorensen2024roadmap}. Decomposing reward into interpretable attributes~\citep{wang2024interpretable,wang2024helpsteer2} surfaces part of this structure, but the distribution over each attribute remains non-degenerate, which a reward model should capture rather than average away.

\paragraph{The Paradigm of Reward Modeling.}
Reward models follow three paradigms. \textbf{Discriminative} RMs regress a scalar over preference pairs with BT loss~\citep{bradley1952rank,stiennon2020learning,ouyang2022training}, scaled through large preference mixtures~\citep{liu2025skywork,cui2023ultrafeedback,he2025air,wang2026helpsteer3} and multi-attribute regression~\citep{wang2024interpretable,wang2024helpsteer2}. \textbf{Generative} RMs emit a textual verdict or critique, from prompting frontier LLMs~\citep{zheng2023judging,gu2024survey} to reasoning-trained and rubric-based judges~\citep{guo2026reward,chen2025rm,gunjal2025rubrics,viswanathan2026checklists}, trading inference cost for interpretability but still producing a scalar score per query. A third line replaces the scalar head with a \textbf{parametric distributional} one: DPL~\citep{siththaranjan2024distributional} and URM~\citep{lou2024uncertainty} predict a Gaussian, DPRM~\citep{li2024aligning} a categorical distribution via optimal transport, and QRM~\citep{dorka2024quantile} a fixed quantile grid, each fixing a distribution family in advance and limiting the shapes $p(\mathbf{r}\mid x,y)$ can take. We instead use a Diffusion Transformer~\citep{peebles2023scalable} as the reward head, representing $p(\mathbf{r}\mid x,y)$ within a single architecture supporting both multi-attribute denoising and a distributional BT objective.
\section{Conclusion}
We presented \textbf{DRM}, a reward model whose DiT head models the conditional reward distribution $p(\mathbf{r}\mid x,y)$ without committing to any parametric family. A single architecture handles both multi-attribute and preference data, and at inference yields distributional statistics and a reward-axis scaling lever unavailable to scalar RMs. Across five benchmarks, DRM matches or surpasses baselines under matched data and backbone, stays competitive with much larger discriminative, distributional, and generative RMs despite its modest training scale and recovers multimodal reward structure that conventional heads cannot. Beyond standard reward-model evaluation, our uncertainty-aware rejection and risk-sensitive aggregation experiments demonstrate that the learned reward distributions provide useful decision signals beyond their mean. We further show that these advantages translate into downstream policy optimization, where DRM achieves improved performance when used as the reward model for RLHF.
\clearpage
\section*{Limitations}
\label{sec:limitations}

This work is an initial exploration of a new reward-modeling paradigm, and several limitations remain. On the one hand, DRM is trained on only a few hundred thousand open-source samples, far below the scale of industrial RMs trained on tens of millions of curated preferences, and does not yet match the strongest open-source scalar RMs in absolute terms. On the other hand, our study fixes a single $8$B encoder and a fixed training-data scale, so how DRM behaves across backbone sizes, model families, and larger data regimes remains unexplored; characterizing these scaling trends is an important direction for establishing the approach.

Moreover, like other reward models, DRM may inherit biases from its training data and could assign high scores to outputs that are stylistically persuasive but factually incorrect or socially harmful. If used as an optimization objective for downstream language models without human oversight or safety constraints, it may amplify such biases or encourage reward hacking behaviors.



\bibliography{drm}

@article{bradley1952rank,
  title={Rank analysis of incomplete block designs: I. the method of paired comparisons},
  author={Bradley, Ralph Allan and Terry, Milton E},
  journal={Biometrika},
  volume={39},
  number={3/4},
  pages={324--345},
  year={1952},
  publisher={JSTOR}
}

@article{christiano2017deep,
  title={Deep reinforcement learning from human preferences},
  author={Christiano, Paul F and Leike, Jan and Brown, Tom and Martic, Miljan and Legg, Shane and Amodei, Dario},
  journal={Advances in neural information processing systems},
  volume={30},
  year={2017}
}

@article{schulman2017proximal,
  title={Proximal policy optimization algorithms},
  author={Schulman, John and Wolski, Filip and Dhariwal, Prafulla and Radford, Alec and Klimov, Oleg},
  journal={arXiv preprint arXiv:1707.06347},
  year={2017}
}

@article{stiennon2020learning,
  title={Learning to summarize with human feedback},
  author={Stiennon, Nisan and Ouyang, Long and Wu, Jeffrey and Ziegler, Daniel and Lowe, Ryan and Voss, Chelsea and Radford, Alec and Amodei, Dario and Christiano, Paul F},
  journal={Advances in neural information processing systems},
  volume={33},
  pages={3008--3021},
  year={2020}
}

@article{song2020score,
  title={Score-based generative modeling through stochastic differential equations},
  author={Song, Yang and Sohl-Dickstein, Jascha and Kingma, Diederik P and Kumar, Abhishek and Ermon, Stefano and Poole, Ben},
  journal={arXiv preprint arXiv:2011.13456},
  year={2020}
}

@article{song2020denoising,
  title={Denoising diffusion implicit models},
  author={Song, Jiaming and Meng, Chenlin and Ermon, Stefano},
  journal={arXiv preprint arXiv:2010.02502},
  year={2020}
}

@article{ho2020denoising,
  title={Denoising diffusion probabilistic models},
  author={Ho, Jonathan and Jain, Ajay and Abbeel, Pieter},
  journal={Advances in neural information processing systems},
  volume={33},
  pages={6840--6851},
  year={2020}
}

@article{dhariwal2021diffusion,
  title={Diffusion models beat gans on image synthesis},
  author={Dhariwal, Prafulla and Nichol, Alexander},
  journal={Advances in neural information processing systems},
  volume={34},
  pages={8780--8794},
  year={2021}
}

@article{lipman2022flow,
  title={Flow matching for generative modeling},
  author={Lipman, Yaron and Chen, Ricky TQ and Ben-Hamu, Heli and Nickel, Maximilian and Le, Matt},
  journal={arXiv preprint arXiv:2210.02747},
  year={2022}
}

@article{bai2022training,
  title={Training a helpful and harmless assistant with reinforcement learning from human feedback},
  author={Bai, Yuntao and Jones, Andy and Ndousse, Kamal and Askell, Amanda and Chen, Anna and DasSarma, Nova and Drain, Dawn and Fort, Stanislav and Ganguli, Deep and Henighan, Tom and others},
  journal={arXiv preprint arXiv:2204.05862},
  year={2022}
}

@article{ho2022classifier,
  title={Classifier-free diffusion guidance},
  author={Ho, Jonathan and Salimans, Tim},
  journal={arXiv preprint arXiv:2207.12598},
  year={2022}
}

@article{ouyang2022training,
  title={Training language models to follow instructions with human feedback},
  author={Ouyang, Long and Wu, Jeffrey and Jiang, Xu and Almeida, Diogo and Wainwright, Carroll and Mishkin, Pamela and Zhang, Chong and Agarwal, Sandhini and Slama, Katarina and Ray, Alex and others},
  journal={Advances in neural information processing systems},
  volume={35},
  pages={27730--27744},
  year={2022}
}

@article{zheng2023judging,
  title={Judging llm-as-a-judge with mt-bench and chatbot arena},
  author={Zheng, Lianmin and Chiang, Wei-Lin and Sheng, Ying and Zhuang, Siyuan and Wu, Zhanghao and Zhuang, Yonghao and Lin, Zi and Li, Zhuohan and Li, Dacheng and Xing, Eric and others},
  journal={Advances in neural information processing systems},
  volume={36},
  pages={46595--46623},
  year={2023}
}

@inproceedings{gao2023scaling,
  title={Scaling laws for reward model overoptimization},
  author={Gao, Leo and Schulman, John and Hilton, Jacob},
  booktitle={International Conference on Machine Learning},
  pages={10835--10866},
  year={2023},
  organization={PMLR}
}

@article{cui2023ultrafeedback,
  title={Ultrafeedback: Boosting language models with scaled ai feedback},
  author={Cui, Ganqu and Yuan, Lifan and Ding, Ning and Yao, Guanming and He, Bingxiang and Zhu, Wei and Ni, Yuan and Xie, Guotong and Xie, Ruobing and Lin, Yankai and others},
  journal={arXiv preprint arXiv:2310.01377},
  year={2023}
}

@inproceedings{peebles2023scalable,
  title={Scalable diffusion models with transformers},
  author={Peebles, William and Xie, Saining},
  booktitle={Proceedings of the IEEE/CVF international conference on computer vision},
  pages={4195--4205},
  year={2023}
}

@article{casper2023open,
  title={Open problems and fundamental limitations of reinforcement learning from human feedback},
  author={Casper, Stephen and Davies, Xander and Shi, Claudia and Gilbert, Thomas Krendl and Scheurer, J{\'e}r{\'e}my and Rando, Javier and Freedman, Rachel and Korbak, Tomasz and Lindner, David and Freire, Pedro and others},
  journal={arXiv preprint arXiv:2307.15217},
  year={2023}
}

@article{rafailov2023direct,
  title={Direct preference optimization: Your language model is secretly a reward model},
  author={Rafailov, Rafael and Sharma, Archit and Mitchell, Eric and Manning, Christopher D and Ermon, Stefano and Finn, Chelsea},
  journal={Advances in neural information processing systems},
  volume={36},
  pages={53728--53741},
  year={2023}
}

@article{dong2023raft,
  title={Raft: Reward ranked finetuning for generative foundation model alignment},
  author={Dong, Hanze and Xiong, Wei and Goyal, Deepanshu and Zhang, Yihan and Chow, Winnie and Pan, Rui and Diao, Shizhe and Zhang, Jipeng and Shum, Kashun and Zhang, Tong},
  journal={arXiv preprint arXiv:2304.06767},
  year={2023}
}

@article{gu2024survey,
  title={A survey on llm-as-a-judge},
  author={Gu, Jiawei and Jiang, Xuhui and Shi, Zhichao and Tan, Hexiang and Zhai, Xuehao and Xu, Chengjin and Li, Wei and Shen, Yinghan and Ma, Shengjie and Liu, Honghao and others},
  journal={The Innovation},
  year={2024},
  publisher={Elsevier}
}

@article{zhang2024diverging,
  title={Diverging preferences: When do annotators disagree and do models know?},
  author={Zhang, Michael JQ and Wang, Zhilin and Hwang, Jena D and Dong, Yi and Delalleau, Olivier and Choi, Yejin and Choi, Eunsol and Ren, Xiang and Pyatkin, Valentina},
  journal={arXiv preprint arXiv:2410.14632},
  year={2024}
}

@article{lambert2024tulu,
  title={Tulu 3: Pushing frontiers in open language model post-training},
  author={Lambert, Nathan and Morrison, Jacob and Pyatkin, Valentina and Huang, Shengyi and Ivison, Hamish and Brahman, Faeze and Miranda, Lester James V and Liu, Alisa and Dziri, Nouha and Lyu, Shane and others},
  journal={arXiv preprint arXiv:2411.15124},
  year={2024}
}

@article{shao2024deepseekmath,
  title={Deepseekmath: Pushing the limits of mathematical reasoning in open language models},
  author={Shao, Zhihong and Wang, Peiyi and Zhu, Qihao and Xu, Runxin and Song, Junxiao and Bi, Xiao and Zhang, Haowei and Zhang, Mingchuan and Li, YK and Wu, Yang and others},
  journal={arXiv preprint arXiv:2402.03300},
  year={2024}
}

@inproceedings{siththaranjan2024distributional,
  title={Distributional preference learning: Understanding and accounting for hidden context in rlhf},
  author={Siththaranjan, Anand and Laidlaw, Cassidy and Hadfield-Menell, Dylan},
  booktitle={International Conference on Learning Representations},
  volume={2024},
  pages={27448--27471},
  year={2024}
}

@inproceedings{wang2024helpsteer,
  title={Helpsteer: Multi-attribute helpfulness dataset for steerlm},
  author={Wang, Zhilin and Dong, Yi and Zeng, Jiaqi and Adams, Virginia and Sreedhar, Makesh Narsimhan and Egert, Daniel and Delalleau, Olivier and Scowcroft, Jane and Kant, Neel and Swope, Aidan and others},
  booktitle={Proceedings of the 2024 Conference of the North American Chapter of the Association for Computational Linguistics: Human Language Technologies (Volume 1: Long Papers)},
  pages={3371--3384},
  year={2024}
}

@article{li2024aligning,
  title={Aligning crowd feedback via distributional preference reward modeling},
  author={Li, Dexun and Zhang, Cong and Dong, Kuicai and Deik, Derrick Goh Xin and Tang, Ruiming and Liu, Yong},
  journal={arXiv preprint arXiv:2402.09764},
  year={2024}
}

@article{wang2024helpsteer2,
  title={Helpsteer 2: Open-source dataset for training top-performing reward models},
  author={Wang, Zhilin and Dong, Yi and Delalleau, Olivier and Zeng, Jiaqi and Shen, Gerald and Egert, Daniel and Zhang, Jimmy J and Sreedhar, Makesh N and Kuchaiev, Oleksii},
  journal={Advances in Neural Information Processing Systems},
  volume={37},
  pages={1474--1501},
  year={2024}
}

@article{sorensen2024roadmap,
  title={A roadmap to pluralistic alignment},
  author={Sorensen, Taylor and Moore, Jared and Fisher, Jillian and Gordon, Mitchell and Mireshghallah, Niloofar and Rytting, Christopher Michael and Ye, Andre and Jiang, Liwei and Lu, Ximing and Dziri, Nouha and others},
  journal={arXiv preprint arXiv:2402.05070},
  year={2024}
}

@article{kirk2024prism,
  title={The PRISM alignment dataset: What participatory, representative and individualised human feedback reveals about the subjective and multicultural alignment of large language models},
  author={Kirk, Hannah R and Whitefield, Alexander and R{\"o}ttger, Paul and Bean, Andrew and Margatina, Katerina and Ciro, Juan and Mosquera, Rafael and Bartolo, Max and Williams, Adina and He, He and others},
  journal={Advances in Neural Information Processing Systems},
  volume={37},
  pages={105236--105344},
  year={2024}
}

@article{wang2024helpsteer2pref,
  title={Helpsteer2-preference: Complementing ratings with preferences},
  author={Wang, Zhilin and Bukharin, Alexander and Delalleau, Olivier and Egert, Daniel and Shen, Gerald and Zeng, Jiaqi and Kuchaiev, Oleksii and Dong, Yi},
  journal={arXiv preprint arXiv:2410.01257},
  year={2024}
}

@article{dorka2024quantile,
  title={Quantile regression for distributional reward models in rlhf},
  author={Dorka, Nicolai},
  journal={arXiv preprint arXiv:2409.10164},
  year={2024}
}

@inproceedings{wang2024interpretable,
  title={Interpretable preferences via multi-objective reward modeling and mixture-of-experts},
  author={Wang, Haoxiang and Xiong, Wei and Xie, Tengyang and Zhao, Han and Zhang, Tong},
  booktitle={Findings of the Association for Computational Linguistics: EMNLP 2024},
  pages={10582--10592},
  year={2024}
}

@article{lou2024uncertainty,
  title={Uncertainty-aware reward model: Teaching reward models to know what is unknown},
  author={Lou, Xingzhou and Yan, Dong and Shen, Wei and Yan, Yuzi and Xie, Jian and Zhang, Junge},
  journal={arXiv preprint arXiv:2410.00847},
  year={2024}
}

@article{mahan2024generative,
  title={Generative reward models},
  author={Mahan, Dakota and Van Phung, Duy and Rafailov, Rafael and Blagden, Chase and Lile, Nathan and Castricato, Louis and Fr{\"a}nken, Jan-Philipp and Finn, Chelsea and Albalak, Alon},
  journal={arXiv preprint arXiv:2410.12832},
  year={2024}
}

@article{gunjal2025rubrics,
  title={Rubrics as rewards: Reinforcement learning beyond verifiable domains},
  author={Gunjal, Anisha and Wang, Anthony and Lau, Elaine and Nath, Vaskar and He, Yunzhong and Liu, Bing and Hendryx, Sean},
  journal={arXiv preprint arXiv:2507.17746},
  year={2025}
}

@article{chen2025rm,
  title={Rm-r1: Reward modeling as reasoning},
  author={Chen, Xiusi and Li, Gaotang and Wang, Ziqi and Jin, Bowen and Qian, Cheng and Wang, Yu and Wang, Hongru and Zhang, Yu and Zhang, Denghui and Zhang, Tong and others},
  journal={arXiv preprint arXiv:2505.02387},
  year={2025}
}

@inproceedings{zhang2025generative,
  title={Generative verifiers: Reward modeling as next-token prediction},
  author={Zhang, Lunjun and Hosseini, Arian and Bansal, Hritik and Kazemi, Seyed Mehran and Kumar, Aviral and Agarwal, Rishabh},
  booktitle={International Conference on Learning Representations},
  volume={2025},
  pages={12476--12505},
  year={2025}
}

@article{guo2025deepseek,
  title={DeepSeek-R1 incentivizes reasoning in LLMs through reinforcement learning},
  author={Guo, Daya and Yang, Dejian and Zhang, Haowei and Song, Junxiao and Wang, Peiyi and Zhu, Qihao and Xu, Runxin and Zhang, Ruoyu and Ma, Shirong and Bi, Xiao and others},
  journal={Nature},
  volume={645},
  number={8081},
  pages={633--638},
  year={2025},
  publisher={Nature Publishing Group UK London}
}

@inproceedings{frick2025evaluate,
  title={How to evaluate reward models for rlhf},
  author={Frick, Evan and Li, Tianle and Chen, Connor and Chiang, Wei-Lin and Angelopoulos, Anastasios and Jiao, Jiantao and Zhu, Banghua and Gonzalez, Joseph E and Stoica, Ion},
  booktitle={International Conference on Learning Representations},
  volume={2025},
  pages={18128--18163},
  year={2025}
}

@inproceedings{liu2025rm,
  title={Rm-bench: Benchmarking reward models of language models with subtlety and style},
  author={Liu, Yantao and Yao, Zijun and Min, Rui and Cao, Yixin and Hou, Lei and Li, Juanzi},
  booktitle={International Conference on Learning Representations},
  volume={2025},
  pages={44323--44355},
  year={2025}
}

@inproceedings{tan2025judgebench,
  title={Judgebench: A benchmark for evaluating llm-based judges},
  author={Tan, Sijun and Zhuang, Siyuan and Montgomery, Kyle and Tang, William and Cuadron, Alejandro and Wang, Chenguang and Popa, Raluca and Stoica, Ion},
  booktitle={International Conference on Learning Representations},
  volume={2025},
  pages={63277--63303},
  year={2025}
}

@article{malik2025rewardbench,
  title={Rewardbench 2: Advancing reward model evaluation},
  author={Malik, Saumya and Pyatkin, Valentina and Land, Sander and Morrison, Jacob and Smith, Noah A and Hajishirzi, Hannaneh and Lambert, Nathan},
  journal={arXiv preprint arXiv:2506.01937},
  year={2025}
}

@inproceedings{zhou2025rmb,
  title={Rmb: Comprehensively benchmarking reward models in llm alignment},
  author={Zhou, Enyu and Zheng, Guodong and Wang, Binghai and Xi, Zhiheng and Dou, Shihan and Bao, Rong and Shen, Wei and Xiong, Limao and Fan, Jessica and Mou, Yurong and others},
  booktitle={International Conference on Learning Representations},
  volume={2025},
  pages={26543--26589},
  year={2025}
}

@article{liu2025skywork,
  title={Skywork-reward-v2: Scaling preference data curation via human-ai synergy},
  author={Liu, Chris Yuhao and Zeng, Liang and Xiao, Yuzhen and He, Jujie and Liu, Jiacai and Wang, Chaojie and Yan, Rui and Shen, Wei and Zhang, Fuxiang and Xu, Jiacheng and others},
  journal={arXiv preprint arXiv:2507.01352},
  year={2025}
}

@article{he2025air,
  title={Air: A systematic analysis of annotations, instructions, and response pairs in preference dataset},
  author={He, Bingxiang and Zhang, Wenbin and Song, Jiaxi and Qian, Cheng and Fu, Zixuan and Sun, Bowen and Ding, Ning and Hong, Haiwen and Huang, Longtao and Xue, Hui and others},
  journal={arXiv preprint arXiv:2504.03612},
  year={2025}
}

@article{viswanathan2026checklists,
  title={Checklists are better than reward models for aligning language models},
  author={Viswanathan, Vijay and Sun, Yanchao and Kong, Xiang and Cao, Meng and Neubig, Graham and Wu, Sherry},
  journal={Advances in Neural Information Processing Systems},
  volume={38},
  pages={114728--114754},
  year={2026}
}

@article{guo2026reward,
  title={Reward reasoning models},
  author={Guo, Jiaxin and Chi, Zewen and Dong, Li and Dong, Qingxiu and Wu, Xun and Huang, Shaohan and Wei, Furu},
  journal={Advances in Neural Information Processing Systems},
  volume={38},
  pages={150477--150510},
  year={2026}
}

@article{wang2026helpsteer3,
  title={Helpsteer3-preference: Open human-annotated preference data across diverse tasks and languages},
  author={Wang, Zhilin and Zeng, Jiaqi and Delalleau, Olivier and Shin, Hoo-Chang and Soares, Felipe and Bukharin, Alexander and Evans, Ellie and Dong, Yi and Kuchaiev, Oleksii},
  journal={Advances in Neural Information Processing Systems},
  volume={38},
  year={2026}
}

@misc{internlm2-7b-reward,
      title={InternLM2 Technical Report},
      author  = {Cai, Zheng and Cao, Maosong and Chen, Haojiong and others},
      year={2024},
      eprint={2403.17297},
      archivePrefix={arXiv},
      primaryClass={cs.CL}
}

@misc{Eurus-RM-7b,
      title={Advancing LLM Reasoning Generalists with Preference Trees}, 
      author={Lifan Yuan and Ganqu Cui and Hanbin Wang and Ning Ding and Xingyao Wang and Jia Deng and Boji Shan and Huimin Chen and Ruobing Xie and Yankai Lin and Zhenghao Liu and Bowen Zhou and Hao Peng and Zhiyuan Liu and Maosong Sun},
      year={2024},
      eprint={2404.02078},
      archivePrefix={arXiv},
}

@misc{Starling-RM-34B,
    title = {Starling-7B: Improving LLM Helpfulness \& Harmlessness with RLAIF},
    url = {},
    author = {Zhu, Banghua and Frick, Evan and Wu, Tianhao and Zhu, Hanlin and Jiao, Jiantao},
    month = {November},
    year = {2023}
}

@misc{Llama-3-OffsetBias-RM-8B,
      title={OffsetBias: Leveraging Debiased Data for Tuning Evaluators},
      author={Junsoo Park and Seungyeon Jwa and Meiying Ren and Daeyoung Kim and Sanghyuk Choi},
      year={2024},
      eprint={2407.06551},
      archivePrefix={arXiv},
      primaryClass={cs.CL}
}

@article{Skywork-Reward-Llama-3.1-8B-v0.2,
  title={Skywork-Reward: Bag of Tricks for Reward Modeling in LLMs},
  author={Liu, Chris Yuhao and Zeng, Liang and Liu, Jiacai and Yan, Rui and He, Jujie and Wang, Chaojie and Yan, Shuicheng and Liu, Yang and Zhou, Yahui},
  journal={arXiv preprint arXiv:2410.18451},
  year={2024}
}

@article{GRM-Llama3-8B-rewardmodel-ft,
  title={Regularizing Hidden States Enables Learning Generalizable Reward Model for LLMs},
  author={Yang, Rui and Ding, Ruomeng and Lin, Yong and Zhang, Huan and Zhang, Tong},
  journal={arXiv preprint arXiv:2406.10216},
  year={2024}
}

@misc{LDL-Reward-Gemma-2-27B-v0.1,
  title        = {{LDL-Reward-Gemma-2-27B-v0.1}},
  author       = {Chen, Shikai and Yuan, Jin and Zhang, Yang and Shi, Zhongchao and Fan, Jianping and Geng, Xin and Rui, Yong},
  year         = {2025},
  url    = {https://huggingface.co/ShikaiChen/LDL-Reward-Gemma-2-27B-v0.1},
  note   = {Label Distribution Learning for Reward Modeling. Tech report forthcoming}
}

@misc{DeepSeek-GRM-27B,
      title={Inference-Time Scaling for Generalist Reward Modeling}, 
      author={Zijun Liu and Peiyi Wang and Runxin Xu and Shirong Ma and Chong Ruan and Peng Li and Yang Liu and Yu Wu},
      year={2025},
      eprint={2504.02495},
      archivePrefix={arXiv},
      primaryClass={cs.CL},
      url={https://arxiv.org/abs/2504.02495}, 
}

@misc{gpt4o,
      title={GPT-4o System Card}, 
      author  = {{OpenAI} and Hurst, Aaron and others},
      year={2024},
      eprint={2410.21276},
      archivePrefix={arXiv},
      primaryClass={cs.CL},
      url={https://arxiv.org/abs/2410.21276}, 
}

@misc{claude35sonnet,
  author = {{Anthropic}},
  title  = {{Claude 3.5 Sonnet}},
  year   = {2024},
  url    = {https://www.anthropic.com/news/claude-3-5-sonnet}
}
\clearpage
\appendix

\section{Relationship to Distribution-Level Preference Likelihood}
\label{app:loss_design}

For a preference pair $(x, y_w, y_l)$, a full distribution-level preference likelihood can be written as: 

$$P_\theta(y_w \succ y_l) = \mathbb{E}_{r_w \sim p_\theta(r|x,y_w),\, r_l \sim p_\theta(r|x,y_l)} \left[ \sigma\left(\frac{r_w-r_l}{\tau}\right) \right]$$

The corresponding distribution-level loss is: 

$$\mathcal{L}_{\text{dist}} = -\log \mathbb{E}_{r_w,r_l} \left[ \sigma\left(\frac{r_w-r_l}{\tau}\right) \right]$$

In contrast, our current objective applies the BT loss to denoised reward estimates recovered during diffusion training: 

$$\mathcal{L}_{\text{point}} = \mathbb{E}_{t,\epsilon} \left[ -\log \sigma\left( \frac{\hat r_{0,w}-\hat r_{0,l}}{\tau} \right) \right] $$

By Jensen's inequality, since $-\log(\cdot)$ is convex, for $Z=\sigma((r_w-r_l)/\tau)$, we have: 

$$-\log \mathbb{E}[Z] \leq \mathbb{E}[-\log Z]$$

Therefore,

$$-\log \mathbb{E}_{r_w,r_l} \left[ \sigma\left(\frac{r_w-r_l}{\tau}\right) \right] \leq \mathbb{E}_{r_w,r_l} \left[ -\log \sigma\left(\frac{r_w-r_l}{\tau}\right) \right]$$

This suggests that applying BT loss to stochastic denoised reward estimates can be viewed as optimizing an upper-bound-style surrogate of the full distribution-level preference loss. Minimizing this surrogate encourages a higher distribution-level win probability $P_\theta(y_w \succ y_l)$, although it is not exactly equivalent to directly optimizing the full distributional likelihood.

\section{Implementation Details of Diffusion Reward Head}
\label{app:dit_details}

This section details the internal structure of the Diffusion Reward Head summarized in \Cref{sec:arch}. The head takes as input a noisy reward $\mathbf{r}_t \in \mathbb{R}^K$, a diffusion timestep $t$, the textual conditioning representation $\mathbf{h} \in \mathbb{R}^{d_{\mathrm{enc}}}$ produced by the frozen backbone, and a reward mask $\mathbf{m} \in \{0, 1\}^K$, and outputs a noise prediction $\hat{\boldsymbol{\epsilon}} \in \mathbb{R}^K$.

\paragraph{Input projection.}
The noisy reward $\mathbf{r}_t$ is first concatenated with the reward mask $\mathbf{m}$ along the feature dimension, then projected into the DiT hidden space through a linear layer:
\[
    \mathbf{z}_r = W_r [\mathbf{r}_t; \mathbf{m}] + \mathbf{b}_r.
\]
A learnable positional embedding $\mathbf{p}$ is added to obtain the input state of the first DiT block:
\[
    \mathbf{z}_0 = \mathbf{z}_r + \mathbf{p}.
\]

\paragraph{Timestep embedding.}
The diffusion timestep $t$ is first encoded through a standard sinusoidal embedding and then passed through a two-layer MLP that projects it to the DiT hidden dimension:
\[
    \mathbf{e}_t = \mathrm{MLP}_t(\mathrm{Sinusoidal}(t)).
\]
This design lets the model identify the current denoising stage and adjust its behavior accordingly across timesteps.

\paragraph{Textual conditioning.}
The textual representation $\mathbf{h}$ is mapped to the same hidden dimension through a separate two-layer MLP:
\[
    \mathbf{e}_h = \mathrm{MLP}_h(\mathbf{h}).
\]
The timestep and textual conditions are then fused additively into a unified conditioning vector:
\[
    \mathbf{c} = \mathbf{e}_t + \mathbf{e}_h.
\]

\paragraph{Conditioning via adaLN-Zero.}
Within each DiT block, the conditioning vector $\mathbf{c}$ is used to generate modulation parameters via adaptive layer normalization (adaLN-Zero)~\citep{peebles2023scalable}, which controls both the self-attention and feed-forward sublayers. This injects timestep and textual information through parameter modulation rather than explicit concatenation, preserving the stability of the DiT architecture. We adopt the adaLN-Zero initialization scheme, in which the modulation and output projection layers are initialized close to zero so that the model begins training near an identity mapping and converges more smoothly.

\paragraph{Output head.}
After the final DiT block, a linear layer maps the hidden representation back to the $K$-dimensional reward space to produce the noise prediction $\hat{\boldsymbol{\epsilon}}$.

\paragraph{Reward Mask.} For DRM-Multi, each training example is annotated on only a subset of the \(K\) reward dimensions. Unannotated dimensions are filled with zeros as placeholders, while the reward mask \(m\) indicates which dimensions are annotated. During training, the target and predicted noise are masked in the denoising objective, such that unannotated dimensions do not contribute to the supervision signal. At inference, we set \(m=\mathbf{1}\), treating all reward dimensions as active and jointly generating the complete reward vector within a single DDIM sampling process.

\section{Benchmark Details}
\label{app:benchmarks}
We describe the five evaluation benchmarks below.
\begin{itemize}[topsep=2pt, partopsep=0pt, leftmargin=12pt, itemsep=0pt]
    \item RewardBench v2. RewardBench v2 is a multi-capability evaluation benchmark for reward models, containing 1,865 test samples. Each sample consists of a prompt, a preferred response, and multiple rejected responses. The benchmark mainly adopts a best-of-4 evaluation format to assess whether a reward model can select the best response. It covers 6 task categories, including factuality, precise instruction following, mathematics, safety, focus, and ties, and is used to evaluate the corresponding capability of reward models.
    \item PPE Preference + PPE Correctness. PPE consists of 2 subsets: PPE Preference and PPE Correctness. PPE Preference contains 16,038 preference pairs. PPE Correctness contains 2,555 prompts, each paired with 32 model responses, and provides correctness labels based on tasks with verifiable answers, including MMLU-Pro, MATH, GPQA, IFEval, and MBPP-Plus. This benchmark supports Best-of-$N$ evaluation and can be used to assess both the model’s alignment with real human preferences and its ability to identify objectively correct responses.
    \item RMB. RMB is a comprehensive benchmark for evaluating reward models, primarily constructed around two alignment objectives: helpfulness and harmlessness. Its samples include both pairwise preference data and Best-of-$N$ data. The significance of RMB lies in the fact that it evaluates not only the pairwise preference judgment ability of reward models, but also their practical capability for candidate selection during inference-time scaling and alignment optimization.
    \item RM-Bench. RM-Bench contains 1,327 test samples, where each sample consists of one prompt, three chosen responses, and three rejected responses. RM-Bench defines three difficulty levels: easy, normal, and hard. The data cover domains including chat, code, math, safety refusal, and safety response. This benchmark primarily evaluates a reward model’s sensitivity to subtle content differences, as well as whether it can still identify genuinely better responses under interference from response length, formatting, and stylistic variation.
    \item JudgeBench. JudgeBench is an evaluation benchmark for LLM judges and reward models, containing 350 response pairs generated by GPT-4o and 270 response pairs generated by Claude-3.5-Sonnet. Each sample consists of a question, two candidate responses, and a preference label based on objective correctness. The data cover tasks such as knowledge, reasoning, mathematics, and coding. Unlike evaluations that primarily rely on subjective human preferences, JudgeBench places greater emphasis on factual and logical correctness, and can therefore be used to assess whether a reward model can identify the objectively more correct response in complex response pairs.
\end{itemize}

\section{Hyperparameter Tuning}
\label{app:hparam}
We tune the architecture and optimization settings of the Diffusion Reward Head and then fix the selected configuration across all benchmarks. The search is conducted on RewardBench v2~\citep{malik2025rewardbench}, with the search space summarized in \Cref{tab:hparam_space}. The tuned dimensions are the DiT hidden size, number of Transformer blocks, number of attention heads, dropout rate, learning rate, batch size, and the diffusion beta schedule; for DRM-Pref-8B we additionally tune the Bradley-Terry loss weight $\lambda_{\mathrm{BT}}$.

\begin{table}[h]
\begin{minipage}[t]{0.36\textwidth}
\centering
\small
\captionof{table}{Hyperparameter search space.}
\resizebox{\linewidth}{!}{%
\begin{tabular}{ll}
\toprule
\textbf{Hyperparameter} & \textbf{Search range} \\
\midrule
Hidden size          & $\{384, 512, 768\}$ \\        
Transformer blocks   & $\{3, 4, 5\}$ \\
Attention heads      & $\{6, 8, 12\}$ \\
Dropout              & $\{0.0, 0.1, 0.2\}$ \\
Learning rate        & $\{0.3, 0.5, 1, 2, 4\}{\times}10^{-4}$ \\
Batch size           & $\{64, 128, 256\}$ \\
Beta schedule        & \{linear, sqcos\} \\
$\lambda_{\mathrm{BT}}$ (Pref only) & $\{0.1, 0.5, 1.0\}$ \\
\bottomrule
\end{tabular}
}

\label{tab:hparam_space}
\end{minipage}\hfill
\begin{minipage}[t]{0.62\textwidth}
\centering
\captionof{table}{Maximum scores for each parameter combination. The final selected core architecture configuration is highlighted. }
\scriptsize
\resizebox{\linewidth}{!}{
\begin{tabular}{ccccc}
\toprule
\textbf{Hidden} & \textbf{Depth} & \textbf{Heads} & \textbf{Beta schedule} & \textbf{RewardBench v2} \\
\midrule
\textbf{384} & \textbf{3} & \textbf{6} & \textbf{sqcos} & \textbf{62.88} \\
384 & 4 & 6 & sqcos & 59.83 \\
384 & 5 & 6 & sqcos & 62.04 \\
512 & 4 & 8 & sqcos & 62.38 \\
512 & 4 & 8 & linear & 61.76 \\
768 & 4 & 12 & sqcos & 54.80 \\
\bottomrule
\end{tabular}
}
\label{tab:hparam_results}
\textit{Inference settings:}
$num\_samples=8$, $guidance\_scale=3.5$, and $num\_steps=20$.
\end{minipage}
\end{table}

Our search was conducted in two stages:
\begin{enumerate}
    \item Core architecture parameter search: We explored the impact of DiT hidden size, number of Transformer blocks, number of attention heads, and the diffusion beta schedule on performance.
    \item Training parameter fine-tuning: With the core architecture fixed, we searched for all feasible combinations of learning rate, dropout, and batch size to optimize convergence speed and training stability.
\end{enumerate}

Table~\ref{tab:hparam_results} presents the maximum RewardBench v2 scores for all core architecture parameter settings across different training parameter combinations. The final selected configuration, 384\_3\_6\_sqcos, is highlighted. The corresponding training parameters are set to lr = 5e-5, Dropout = 0.2, and Batch = 64. This model achieves the best performance among all combinations while maintaining low training and inference cost, which is the final DRM-Multi-8B model.

After determining the core architecture parameters, we fixed the structure and trained DRM-Pref-8B with different values of $\lambda_{BT}$, The value $\lambda_{BT}$ =0.5, which achieved the best performance on the validation set, was ultimately selected for the DRM-Pref-8B model.

\begin{table*}[]
    \centering
    \caption{BoN split of RMB results of different reward models.}
    \label{tab:RMB}
    \resizebox{\textwidth}{!}{
    \begin{tabular}{lccc}
    \toprule
        Model &  Helpfulness (BoN) &  Harmlessness (BoN) & Avg. \\ 
        \midrule
        ArmoRM-Llama3-8B-v0.1 & 63.6  & 49.7  & 56.7  \\ 
        Skywork-Reward-Llama-3.1-8B-v0.2 & 60.5  & 56.8  & 58.7  \\ 
        internlm2-7b-reward & 62.6  & 56.3  & 59.5  \\ 
        DeepSeek-GRM-27B & 63.9  & 58.0  & 61.0  \\ 
        Eurus-RM-7b & 67.9  & 54.3  & 61.1  \\ 
        Claude-3.5-Sonnet-20240620 & 70.5  & 51.8  & 61.2  \\ 
        Skywork-Reward-Gemma-2-27B-v0.2 & 63.1  & 59.9  & 61.5  \\ 
        GPT-4o-20240513 & 63.9  & 68.2  & 66.1  \\ 
        \midrule
        DRM-Multi-8B & 66.2  & 61.4  & 63.8  \\ 
        DRM-Pref-8B & 66.5  & 61.3  & 63.9 \\ 
        \bottomrule
    \end{tabular}}
    \vspace{-8pt}
\end{table*}

\begin{table*}[]
    \centering
    \caption{JudgeBench results of different reward models.}
    \label{tab:JudgeBench}
    \resizebox{\textwidth}{!}{
    \begin{tabular}{lccccc}
    \toprule
        Model & Knowledge & Reasoning & Math & Code & Avg.  \\ 
        \midrule
        Arena-Hard/GPT-4o-20240513 & 51.6  & 52.6  & 68.4  & 49.1  & 54.3   \\ 
        Arena-Hard/Claude-3.5-Sonnet-20240620 & 52.3  & 59.6  & 61.0  & 48.3  & 54.6   \\ 
        \midrule
        DRM-Multi-8B & 55.8  & 62.0  & 63.7  & 57.9  & 58.6   \\ 
        DRM-Pref-8B & 54.9  & 57.0  & 62.2  & 60.3  & 57.1  \\ \bottomrule
    \end{tabular}}
    \vspace{-8pt}
\end{table*}

\begin{table*}[t]
    \centering
    
    \caption{PPE correctness results of different reward models.}
    \label{tab:PPE_Corr}
    \resizebox{\textwidth}{!}{
    \begin{tabular}{lcccccc}
    \toprule
        Model & PPE-MMLU & PPE-MATH & PPE-GPQA & PPE-IFEval & PPE-MBPP & Avg.  \\
        \midrule
        Skywork-Reward-Gemma-2-27B & 53.9  & 62.7  & 52.7  & 53.8  & 59.3  & 56.5 \\  
        Eurus-RM-7b & 63.4  & 69.3  & 53.9  & 59.4  & 54.0  & 60.0  \\ 
        Starling-RM-34B & 67.7  & 66.4  & 57.0  & 56.1  & 54.6  & 60.3   \\
        internlm2-7b-reward & 66.7  & 72.6  & 54.6  & 64.0  & 43.9  & 60.4   \\
        Skywork-Reward-Llama-3.1-8B-v0.2 & 64.3  & 69.6  & 56.5  & 61.5  & 51.6  & 60.7   \\
        ArmoRM-Llama3-8B-v0.1 & 66.5  & 70.7  & 57.0  & 58.4  & 54.2  & 61.4   \\ \midrule
        DRM-Multi-8B & 66.1  & 69.9  & 57.5  & 60.1  & 65.3  & 63.8\\ 
        DRM-Pref-8B & 66.3  & 69.1  & 56.2  & 61.5  & 59.3  & 62.5  \\ 
        \bottomrule
    \end{tabular}}
    \vspace{-8pt}
\end{table*}

\section{More Ablation Studies}

We add further ablation studies. Unless otherwise specified, we keep all other training and inference configurations identical to those used in the main experiments and evaluate performance on RewardBench v2. The default inference configuration uses \(S=10\) DDIM sampling steps, a guidance scale of \(\omega=7\), and \(N=32\) reward samples.

\subsection{DDIM Sampling Steps and Guidance Scale}

\vspace{+3pt}\noindent\textbf{DDIM Sampling Steps} We evaluate \(S \in \{5,10,50,100\}\), with the results shown in Table~\ref{tab:ablation_steps}. For both DRM variants, \(S=10\) achieves the best performance. Reducing the number of sampling steps from 10 to 5 results in only a modest performance drop, while increasing the number of steps provides no further benefit and can even substantially degrade reward-ranking performance. These results suggest that, unlike high-dimensional image generation, DRM requires only a small number of sampling steps to obtain effective reward estimates. We therefore use \(S=10\) as the default setting.

\begin{table}[h]
\begin{minipage}[t]{0.48\textwidth}
\centering
\small
\setlength{\tabcolsep}{8pt}
\caption{Ablation on the number of DDIM sampling steps. We fix the guidance scale to $\omega=7$ and the number of reward samples to $N=32$.}
\label{tab:ablation_steps}
\begin{tabular}{lcc}
\toprule
Model & DDIM Steps $S$ & RewardBench v2 \\
\midrule
DRM-Multi & 5   & 64.9 \\
DRM-Multi & \textbf{10}  & \textbf{65.6} \\
DRM-Multi & 50  & 51.3 \\
DRM-Multi & 100 & 49.4 \\
\midrule
DRM-Pref  & 5   & 64.0 \\
DRM-Pref  & \textbf{10}  & \textbf{65.7} \\
DRM-Pref  & 50  & 55.7 \\
DRM-Pref  & 100 & 61.0 \\
\bottomrule
\end{tabular}

\end{minipage}\hfill
\begin{minipage}[t]{0.48\textwidth}
\centering
\small
\setlength{\tabcolsep}{8pt}
\caption{Ablation on the classifier-free guidance scale. We fix the number of DDIM sampling steps to $S=10$ and the number of samples to $N=32$.}
\label{tab:ablation_guidance}
\begin{tabular}{lcc}
\toprule
Model & Guidance $\omega$ & RewardBench v2 \\
\midrule
DRM-Multi & 1    & 57.6 \\
DRM-Multi & 3.5  & 64.1 \\
DRM-Multi & \textbf{7}    & \textbf{65.6} \\
DRM-Multi & 14   & 57.6 \\
\midrule
DRM-Pref  & 1    & 61.8 \\
DRM-Pref  & 3.5  & 65.5 \\
DRM-Pref  & \textbf{7}    & \textbf{65.7} \\
DRM-Pref  & 14   & 62.4 \\
\bottomrule
\end{tabular}

\end{minipage}
\end{table}

\vspace{+3pt}\noindent\textbf{Guidance Scale} We evaluate \(\omega \in \{1, 3.5, 7, 14\}\), with the results shown in Table~\ref{tab:ablation_guidance}. Both DRM variants exhibit a clear non-monotonic trend. Weak guidance does not sufficiently leverage the textual condition, whereas overly strong guidance also degrades reward-ranking performance. A moderate guidance scale of \(\omega=7\) achieves the best results for both DRM-Multi and DRM-Pref, and is therefore adopted as the default setting.

\subsection{Number of Reward Samples}

\begin{wraptable}{r}{0.50\textwidth}
\centering
\small
\caption{Ablation on the reward mask used during inference for DRM-Multi.}
\label{tab:ablation_mask}
\begin{tabular}{lc}
\toprule
Inference Mask & RewardBench v2 \\
\midrule
Training masks & 65.1 \\
All-one mask            & \textbf{65.6} \\
\bottomrule
\end{tabular}
\end{wraptable}

At inference time, DRM constructs an empirical reward distribution by drawing multiple samples from the learned conditional reward distribution. We therefore further study the effect of the number of reward samples \(N\) on reward estimation. We fix \(S=10\) and \(\omega=7\), and gradually increase \(N\) from 1 to 32. As shown in Figure~\ref{fig:tts-b}, DRM-Multi improves consistently from 56.5 on RewardBench v2 with \(N=1\) to 65.6 with \(N=32\), while DRM-Pref improves from 64.4 to 65.7. These results indicate that using more reward samples provides a more stable empirical approximation of the learned conditional reward distribution and reduces the influence of randomness from individual samples on the final reward estimate. We therefore use \(N=32\) as the default setting.

\subsection{Reward Mask}

DRM-Multi is trained in a unified 19-dimensional reward space, while each training example typically contains annotations for only a subset of the reward dimensions. During training, we use a reward mask to distinguish observed labels from missing dimensions and compute the denoising loss only over the annotated dimensions.

At inference time, we compare two masking strategies. The first follows the attribute structure of the training data and generates the complete reward vector through multiple sampling processes with the corresponding training masks. The second uses an all-one mask, treating all reward dimensions as valid and jointly generating the full reward vector within a single sampling process.

As shown in Table~\ref{tab:ablation_mask}, using the all-one mask improves the RewardBench v2 score from 65.1 to 65.6. This result indicates that jointly generating all reward dimensions at inference time does not degrade performance and can better exploit the cross-attribute structure learned from heterogeneous multi-attribute supervision. We therefore use the all-one mask in all DRM-Multi experiments, jointly generating the complete 19-dimensional reward vector within a single DDIM sampling process.

\subsection{Reward Dimension}

One important feature of DRM-Multi is its ability to integrate heterogeneous multi-attribute supervision from multiple data sources within a unified reward space. To study the effect of reward-space breadth on model performance, we further train a 5-dimensional DRM variant on the largest single UltraFeedback subset and compare it with the 19-dimensional DRM-Multi-Half model trained at a comparable data scale. The results are shown in Table~\ref{tab:ablation_reward_dim}.

\begin{table}[t]
\centering
\small
\setlength{\tabcolsep}{7pt}
\caption{Analysis of reward-space dimensionality under comparable training scales.}
\label{tab:ablation_reward_dim}
\begin{tabular}{lccc}
\toprule
Model & Reward Dim. & Training Samples & RewardBench v2 \\
\midrule
UltraFeedback variant & 5  & 240K & 54.6 \\
DRM-Multi-Half        & 19 & 284K & \textbf{63.9} \\
\bottomrule
\end{tabular}
\end{table}

Under similar training scales, the 19-dimensional DRM-Multi-Half achieves a RewardBench v2 score of 63.9, substantially outperforming the 5-dimensional UltraFeedback variant at 54.6. This result suggests that integrating attribute-level supervision from multiple sources into a richer unified reward space provides DRM with more informative reward-modeling signals.

\subsection{Training Scale and Supervision Regime}
\label{sec:training_scale}

DRM uses the same diffusion reward framework to support two forms of supervision: DRM-Multi is trained with multi-attribute reward annotations, while DRM-Pref is trained with pairwise preference data. Since the full DRM-Multi model uses 569K training examples whereas DRM-Pref uses 273K preference pairs, a direct comparison between the two confounds supervision type with training scale. To better disentangle these factors, we construct DRM-Multi-Half by stratified sampling according to data source and retaining 50\% of the DRM-Multi training corpus. This yields a training scale closer to that of DRM-Pref while keeping the model architecture and inference configuration unchanged. The results are shown in Table~\ref{tab:ablation_supervision}.

Reducing the DRM-Multi training data by half decreases its performance from 66.2 to 65.1, showing that training scale has a clear impact on DRM performance. More importantly, under comparable training scales, DRM-Pref achieves an average score of 65.8, outperforming DRM-Multi-Half at 65.1. This indicates that the diffusion reward framework can effectively leverage pairwise preference supervision, and that the performance gap between the full DRM-Multi and DRM-Pref is largely attributable to the difference in training scale rather than an inherent weakness of pairwise supervision.

The two supervision regimes also exhibit complementary strengths across benchmarks. DRM-Pref performs slightly better on preference-oriented benchmarks, achieving 65.7 on RewardBench v2, 63.0 on PPE Preference, and 78.2 on RMB Pairwise, all slightly higher than DRM-Multi. In contrast, the full DRM-Multi performs better on correctness-oriented benchmarks, reaching 63.8 on PPE Correctness and 58.6 on JudgeBench. Overall, pairwise preference supervision more directly strengthens relative preference judgments, whereas the richer attribute-level signals provided by multi-attribute supervision support a more comprehensive reward representation.

These results further demonstrate that DRM does not depend on a particular form of reward supervision, but remains effective and competitive under both multi-attribute reward learning and pairwise preference learning.

\begin{table}[t]
\centering
\small
\caption{Comparison of DRM variants under different training scales and supervision regimes. DRM-Multi-Half is trained on a stratified 50\% subset of the DRM-Multi training corpus.}
\label{tab:ablation_supervision}
\resizebox{\linewidth}{!}{
\begin{tabular}{llcccccccc}
\toprule
Model & Supervision & Training Samples & RewardBench v2 & PPE Pref & PPE Corr & RMB Pairwise & RM-Bench & JudgeBench & Avg. \\
\midrule
DRM-Multi      & Multi-attribute     & 569K       & 65.6 & 62.5 & \textbf{63.8} & 78.0          & \textbf{68.8} & \textbf{58.6} & \textbf{66.2} \\
DRM-Multi-Half & Multi-attribute     & 284K & 63.9 & 61.3 & 63.2          & 77.0          & 68.6          & 56.8          & 65.1          \\
DRM-Pref       & Pairwise preference & 273K      & \textbf{65.7} & \textbf{63.0} & 62.5 & \textbf{78.2} & 68.1 & 57.1 & 65.8 \\
\bottomrule
\end{tabular}
}
\end{table}

\subsection{Stability Across Inference Seeds}

To evaluate the robustness of DRM to stochastic diffusion sampling, we fix the trained checkpoints and all inference hyperparameters, and vary only the inference random seed. We evaluate both DRM-Multi-8B and DRM-Pref-8B using 10 seeds, \(\{0,\ldots,9\}\), with \(S=10\) DDIM steps, \(N=32\) reward samples, and guidance scale \(\omega=7\), covering all benchmarks in Table~\ref{tab:main_results}. ArmoRM is deterministic in evaluation mode, so we report its single-run score as a fixed reference. For DRM, we report the mean, standard deviation, and 95\% confidence interval across the 10 inference seeds.

\begin{table}[t]
\centering
\caption{Stability of DRM across 10 inference seeds. We report the mean, standard deviation, and 95\% confidence interval over seeds \(\{0,\ldots,9\}\). All DRM evaluations use \(S=10\), \(N=32\), and \(\omega=7\). ArmoRM is deterministic in evaluation mode and is reported as a fixed single-run reference.}
\label{tab:seed_stability}
\resizebox{\linewidth}{!}{
\begin{tabular}{lccccc}
\toprule
\multirow{2}{*}{Benchmark} &
\multicolumn{2}{c}{DRM-Multi-8B} &
\multicolumn{2}{c}{DRM-Pref-8B} &
\multirow{2}{*}{ArmoRM} \\
\cmidrule(lr){2-3} \cmidrule(lr){4-5}
& Mean $\pm$ Std & 95\% CI
& Mean $\pm$ Std & 95\% CI
& \\
\midrule
RewardBench v2 & $65.133 \pm 0.519$ & $[64.761, 65.504]$ & $65.655 \pm 0.197$ & $[65.514, 65.796]$ & $66.5$ \\
PPE Pref       & $62.047 \pm 0.138$ & $[61.948, 62.146]$ & $63.047 \pm 0.133$ & $[62.952, 63.142]$ & $60.6$ \\
PPE Corr       & $63.653 \pm 0.103$ & $[63.580, 63.727]$ & $62.446 \pm 0.078$ & $[62.391, 62.502]$ & $61.4$ \\
RMB            & $77.966 \pm 0.071$ & $[77.915, 78.017]$ & $78.255 \pm 0.042$ & $[78.225, 78.285]$ & $64.6$ \\
RM-Bench       & $68.917 \pm 0.156$ & $[68.806, 69.029]$ & $68.060 \pm 0.093$ & $[67.994, 68.127]$ & $67.7$ \\
JudgeBench     & $57.845 \pm 0.569$ & $[57.438, 58.252]$ & $57.161 \pm 0.368$ & $[56.898, 57.425]$ & $53.2$ \\
\midrule
\textbf{Avg-6} & $\mathbf{65.927 \pm 0.183}$ & $\mathbf{[65.796, 66.058]}$ & $\mathbf{65.771 \pm 0.100}$ & $\mathbf{[65.700, 65.842]}$ & $\mathbf{62.3}$ \\
\bottomrule
\end{tabular}
}
\end{table}

As shown in Table~\ref{tab:seed_stability}, DRM remains highly stable across different inference seeds. DRM-Multi-8B achieves an Avg-6 score of \(65.927 \pm 0.183\), with a 95\% confidence interval of \([65.796, 66.058]\), while DRM-Pref-8B achieves \(65.771 \pm 0.100\), with a 95\% confidence interval of \([65.700, 65.842]\). Both substantially outperform the deterministic ArmoRM score of \(62.300\). The improvements of \(+3.627\) and \(+3.471\), respectively, are much larger than the variation induced by stochastic diffusion sampling, showing that DRM's performance gains are robust to inference randomness.

\section{Inference Efficiency}

We conducted an experiment of inference-efficiency analysis. We compare DRM with the scalar reward model FsfairX and the single-pass distributional reward model QRM. We further examine how the number of reward samples \(N\) and the number of DDIM sampling steps \(S\) affect inference latency, and study the accuracy-latency trade-off obtained by reducing the number of denoising steps.

\subsection{Measurement Setup}

We randomly sample 256 prompt-response examples from RewardBench v2 and evaluate all models with batch size 1. For DRM, we use guidance scale $\omega=7$, reward sample counts $N \in \{1,8,32\}$, and DDIM steps $S \in \{5, 10\}$. We use \(S=10\) and \(N=32\) as the default inference configuration throughout the main experiments.

We separately measure the latency of the frozen text encoder, the RewardDiT head, and the complete end-to-end scoring pipeline. End-to-end latency includes text encoding, DDIM sampling, reward aggregation, and other necessary scheduling and data-transfer overhead, and therefore does not necessarily equal the simple sum of encoder and RewardDiT latency. We additionally report sample throughput, token throughput, and peak GPU memory.

\subsection{Results}

\begin{table*}[t]
\centering
\small
\caption{Inference efficiency of DRM and baseline reward models. Latency is measured with batch size 1 on 256 randomly sampled RewardBench v2 examples. For DRM, $\omega=7$. Relative cost is normalized by the end-to-end latency of FsfairX.}
\label{tab:inference_efficiency}
\resizebox{\textwidth}{!}{
\begin{tabular}{lcccccccc}
\toprule
Model & $N$ & $S$ & Encoder Lat. (ms) & RewardDiT Lat. (ms) & E2E Lat. (ms) & Throughput (samples/s) & Peak Mem. (GB) & Rel. Cost \\
\midrule
FsfairX & -- & -- & 38.983 & -- & 38.983 & 25.652 & 14.104 & $1.000\times$ \\
QRM & -- & -- & 48.099 & -- & 48.099 & 20.791 & 14.121 & $1.234\times$ \\
\midrule
DRM & 1 & 10 & 38.834 & 21.197 & 62.403 & 16.025 & 14.152 & $1.601\times$ \\
DRM & 8 & 10 & 38.834 & 21.799 & 63.259 & 15.808 & 14.152 & $1.623\times$ \\
DRM & 32 & 10 & 38.834 & 22.461 & 63.559 & 15.734 & 14.152 & $1.630\times$ \\
\midrule
DRM & 1 & 5 & 41.246 & 10.501 & 53.231 & 18.786 & 14.156 & $1.365\times$ \\
DRM & 8 & 5 & 41.246 & 10.840 & 53.544 & 18.676 & 14.156 & $1.373\times$ \\
DRM & 32 & 5 & 41.246 & 11.054 & 53.755 & 18.603 & 14.156 & $1.378\times$ \\
\bottomrule
\end{tabular}
}
\vspace{2pt}
\end{table*}

Table~\ref{tab:inference_efficiency} summarizes the inference cost of DRM. Under the default \(N=32,S=10\) setting, DRM requires 63.559 ms end-to-end latency, corresponding to \(1.63\times\) the cost of FsfairX, while peak memory increases by only 0.048 GB. The additional overhead mainly comes from the lightweight RewardDiT denoising process.

Importantly, latency grows only slightly with the number of reward samples because samples are processed in parallel: at \(S=10\), increasing \(N\) from 1 to 32 raises RewardDiT latency from 21.197 ms to 22.461 ms. In contrast, reducing DDIM steps from \(S=10\) to \(S=5\) nearly halves RewardDiT latency from 22.461 ms to 11.054 ms, showing that \(S\) is the primary control knob for inference cost.

\subsection{Accuracy-Latency Trade-off}

\begin{table*}[t]
\centering
\small
\caption{Reward-model performance under different DDIM sampling budgets. All DRM results use $N=32$ and guidance scale $\omega=7$. The measured end-to-end latency is 63.559 ms for $S=10$ and 53.755 ms for $S=5$, corresponding to $1.630\times$ and $1.378\times$ the latency of FsfairX, respectively.}
\label{tab:ddim_performance}
\resizebox{\textwidth}{!}{
\begin{tabular}{lcccccccc}
\toprule
Model & $S$ & RBv2 & PPE Pref. & PPE Corr. & RMB & RM-Bench & JudgeBench & Avg. \\
\midrule
DRM-Multi & 10 & 65.6 & 62.5 & 63.8 & 78.0 & 68.8 & 58.6 & \textbf{66.2} \\
DRM-Multi & 5 & 64.9 & 62.1 & 63.9 & 77.7 & 69.0 & 57.4 & 65.9 \\
\midrule
DRM-Pref & 10 & 65.7 & 63.0 & 62.5 & 78.2 & 68.1 & 57.1 & \textbf{65.8} \\
DRM-Pref & 5 & 64.0 & 62.6 & 60.8 & 77.3 & 67.1 & 57.5 & 64.9 \\
\bottomrule
\end{tabular}
}
\end{table*}

We next evaluate the accuracy-latency trade-off of reducing DDIM steps. As shown in Table~\ref{tab:ddim_performance}, decreasing \(S\) from 10 to 5 reduces the \(N=32\) end-to-end latency from 63.559 ms to 53.755 ms (15.4\%), lowering the relative cost from \(1.63\times\) to \(1.38\times\). DRM-Multi remains stable, with its average score decreasing only from 66.2 to 65.9, while DRM-Pref drops from 65.8 to 64.9.

The \(S=5\) setting also approaches the inference cost of QRM (53.755 vs. 48.099 ms), while DRM-Multi still outperforms QRM in average score (65.9 vs. 64.1). We therefore use \(S=10\) by default and provide \(S=5\) as a lower-latency alternative.

\subsection{Offline Encoder Caching}

Because DRM uses a frozen FsfairX encoder, the textual representation \(h=\mathrm{Enc}(x,y)\) can be precomputed and reused. For the 256 RewardBench v2 samples, embedding construction takes 9.385 s in total, or 36.662 ms per sample, and requires only 4.360 MB of storage. In repeated scoring settings such as benchmark evaluation or fixed candidate-pool ranking, subsequent inference therefore only runs the lightweight RewardDiT head, which takes about 22 ms per example under the default \(S=10,N=32\) setting.

Overall, DRM provides a tunable inference budget: the default configuration costs about 1.63\(\times\) the latency of FsfairX, while \(S=5\) reduces this to 1.38\(\times\). Increasing \(N\) adds little wall-clock overhead because reward samples are processed in parallel.
\section{Other Experiments Details}

We provide per-task breakdowns on RMB, JudgeBench, and PPE Correctness in \Cref{tab:RMB}, \Cref{tab:JudgeBench}, and \Cref{tab:PPE_Corr}. The RMB results use its Best-of-$N$ splits and are not directly comparable to the RMB pairwise scores in \Cref{tab:main_results}.

The full scoring pipeline consists of a frozen FsfairX-LLaMA3-RM-v0.1 encoder and a RewardDiT denoiser. The FsfairX encoder has 7.50B parameters, while RewardDiT has approximately 12.0M parameters; therefore, the full encoder-plus-denoiser pipeline contains approximately 7.52B parameters. Only the RewardDiT component is trained.

\label{sec:rlhf_details}
We conduct downstream RLHF experiments using PPO implemented in veRL, with full-parameter optimization of the actor. All experiments start from the same allenai/Llama-3.1-Tulu-3-8B-SFT initialization and use prompts from UltraFeedback. The original prompt set contains 63,967 examples. After filtering overly long prompts and discarding the final incomplete batch, 62,880 prompts are used for each run. We train for one epoch with a rollout batch size of 480, resulting in 131 rollout-update steps. Each rollout batch is optimized for one PPO epoch with a minibatch size of 120. The actor and critic learning rates are \(1\times10^{-6}\) and \(1\times10^{-5}\), respectively. We use a PPO clip ratio of 0.2, a value-function clip of 0.5, and GAE with \(\gamma=1.0\) and \(\lambda=1.0\). For reward computation, FsfairX and ArmoRM each provide a single scalar reward. DRM-Multi produces 32 reward samples over 19 reward dimensions; we first average over the reward samples and then average across dimensions to obtain the scalar reward used by PPO. The resulting policies are evaluated on Arena-Hard v2 and MT-Bench using GPT-4.1 as the evaluator.

Experiments were run on NVIDIA A800-SXM4-80GB GPUs using PyTorch 2.7.0+cu126, CUDA 12.6, Hugging Face Transformers/Datasets, and Diffusers. RewardDiT checkpoint training required approximately 1.54 GPU-hours in total, including per-epoch validation: 1.11 GPU-hours for the ArmoRM checkpoint and 0.43 GPU-hours for the Tulu3 pair-preference checkpoint. These estimates exclude the one-time cost of generating text embeddings with the frozen FsfairX encoder.

\section{Responsible NLP and Artifact Use}
\label{sec:responsible-nlp}

\paragraph{Artifact Use and Access Conditions.}
We use publicly available datasets, benchmarks, pretrained models, and baseline models for research, training, evaluation, and comparison purposes. We cite the original creators of these artifacts and use them in accordance with their respective licenses, terms of use, and access conditions. We do not redistribute artifacts whose licenses or access conditions prohibit redistribution.

\paragraph{Data Privacy and Sensitive Content.}
We do not collect new personal data or attempt to identify individuals. Our experiments use publicly available datasets and benchmarks. Some preference, safety, or harmlessness-oriented datasets may contain sensitive, toxic, or offensive content as part of their intended research and evaluation scope. We use such data only for research and aggregate evaluation.

\paragraph{LLM Usage.}
During the preparation of this manuscript, we used large language models (LLMs) to assist with language polishing and improving the clarity and readability of the paper. The LLMs were not used to generate research hypotheses, design the methodology, conduct experiments, analyze results, or draw conclusions. All LLM-assisted edits were carefully reviewed and revised by the authors, who take full responsibility for the final content of the manuscript.

\end{document}